\PassOptionsToPackage{unicode}{hyperref}
\PassOptionsToPackage{hyphens}{url}
\PassOptionsToPackage{dvipsnames,svgnames,x11names}{xcolor}
\documentclass[
  12pt,
  letterpaper,
]{article}
\usepackage{xcolor}
\usepackage[margin=1in]{geometry}
\usepackage{amsmath,amssymb}
\usepackage{iftex}
\ifPDFTeX
  \usepackage[T1]{fontenc}
  \usepackage[utf8]{inputenc}
  \usepackage{textcomp} 
\else 
  \usepackage{unicode-math} 
  \defaultfontfeatures{Scale=MatchLowercase}
  \defaultfontfeatures[\rmfamily]{Ligatures=TeX,Scale=1}
\fi
\usepackage{lmodern}
\ifPDFTeX\else
\fi
\IfFileExists{upquote.sty}{\usepackage{upquote}}{}
\IfFileExists{microtype.sty}{
  \usepackage[]{microtype}
  \UseMicrotypeSet[protrusion]{basicmath} 
}{}
\makeatletter
\@ifundefined{KOMAClassName}{
  \IfFileExists{parskip.sty}{%
    \usepackage{parskip}
  }{
    \setlength{\parindent}{0pt}
    \setlength{\parskip}{6pt plus 2pt minus 1pt}}
}{
  \KOMAoptions{parskip=half}}
\makeatother
\makeatletter
\ifx\paragraph\undefined\else
  \let\oldparagraph\paragraph
  \renewcommand{\paragraph}{
    \@ifstar
      \xxxParagraphStar
      \xxxParagraphNoStar
  }
  \newcommand{\xxxParagraphStar}[1]{\oldparagraph*{#1}\mbox{}}
  \newcommand{\xxxParagraphNoStar}[1]{\oldparagraph{#1}\mbox{}}
\fi
\ifx\subparagraph\undefined\else
  \let\oldsubparagraph\subparagraph
  \renewcommand{\subparagraph}{
    \@ifstar
      \xxxSubParagraphStar
      \xxxSubParagraphNoStar
  }
  \newcommand{\xxxSubParagraphStar}[1]{\oldsubparagraph*{#1}\mbox{}}
  \newcommand{\xxxSubParagraphNoStar}[1]{\oldsubparagraph{#1}\mbox{}}
\fi
\makeatother

\usepackage{longtable,booktabs,array}
\usepackage{calc} 
\usepackage{etoolbox}
\makeatletter
\patchcmd\longtable{\par}{\if@noskipsec\mbox{}\fi\par}{}{}
\makeatother
\IfFileExists{footnotehyper.sty}{\usepackage{footnotehyper}}{\usepackage{footnote}}
\makesavenoteenv{longtable}
\usepackage{graphicx}
\makeatletter
\newsavebox\pandoc@box
\newcommand*\pandocbounded[1]{
  \sbox\pandoc@box{#1}%
  \Gscale@div\@tempa{\textheight}{\dimexpr\ht\pandoc@box+\dp\pandoc@box\relax}%
  \Gscale@div\@tempb{\linewidth}{\wd\pandoc@box}%
  \ifdim\@tempb\p@<\@tempa\p@\let\@tempa\@tempb\fi
  \ifdim\@tempa\p@<\p@\scalebox{\@tempa}{\usebox\pandoc@box}%
  \else\usebox{\pandoc@box}%
  \fi%
}
\def\fps@figure{htbp}
\makeatother

\NewDocumentCommand\citeproctext{}{}
\NewDocumentCommand\citeproc{mm}{%
  \begingroup\def\citeproctext{#2}\cite{#1}\endgroup}
\makeatletter
 \let\@cite@ofmt\@firstofone
 \def\@biblabel#1{}
 \def\@cite#1#2{{#1\if@tempswa , #2\fi}}
\makeatother
\newlength{\cslhangindent}
\newlength{\csllabelwidth}
\newenvironment{CSLReferences}[2] 
 {\begin{list}{}{%
  \setlength{\itemindent}{0pt}
  \setlength{\leftmargin}{0pt}
  \setlength{\parsep}{0pt}
  \ifodd #1
   \setlength{\leftmargin}{\cslhangindent}
   \setlength{\itemindent}{-1\cslhangindent}
  \fi
  \setlength{\itemsep}{#2\baselineskip}}}
 {\end{list}}
\usepackage{calc}

\providecommand{\tightlist}{%
  \setlength{\itemsep}{0pt}\setlength{\parskip}{0pt}}

\usepackage{setspace}
\usepackage{booktabs}
\usepackage{array}
\usepackage{tabularx}
\usepackage{longtable}
\usepackage{threeparttable}
\usepackage{caption}
\usepackage{needspace}
\usepackage{float}
\makeatletter
\def\fps@table{H}
\makeatother
\usepackage{etoolbox}
\AtBeginEnvironment{abstract}{\singlespacing}
\AtBeginEnvironment{longtable}{\singlespacing}
\makeatletter
\@ifpackageloaded{caption}{}{\usepackage{caption}}
\AtBeginDocument{%
\ifdefined\contentsname
  \renewcommand*\contentsname{Table of contents}
\else
  \newcommand\contentsname{Table of contents}
\fi
\ifdefined\listfigurename
  \renewcommand*\listfigurename{List of Figures}
\else
  \newcommand\listfigurename{List of Figures}
\fi
\ifdefined\listtablename
  \renewcommand*\listtablename{List of Tables}
\else
  \newcommand\listtablename{List of Tables}
\fi
\ifdefined\figurename
  \renewcommand*\figurename{Figure}
\else
  \newcommand\figurename{Figure}
\fi
\ifdefined\tablename
  \renewcommand*\tablename{Table}
\else
  \newcommand\tablename{Table}
\fi
}
\@ifpackageloaded{float}{}{\usepackage{float}}
\floatstyle{ruled}
\@ifundefined{c@chapter}{\newfloat{codelisting}{h}{lop}}{\newfloat{codelisting}{h}{lop}[chapter]}
\floatname{codelisting}{Listing}

\makeatother
\makeatletter
\@ifpackageloaded{caption}{}{\usepackage{caption}}
\@ifpackageloaded{subcaption}{}{\usepackage{subcaption}}
\makeatother
\usepackage{bookmark}
\IfFileExists{xurl.sty}{\usepackage{xurl}}{} 
\makeatletter
\@ifundefined{xmpquote}{}{}
\makeatother
\hypersetup{
  pdftitle={Toward Embedding-Based Psychometrics: Structural Modeling of Assessment-Item Semantics With Contextual Scores},
  pdfauthor={Jinsong Chen and Shi-Ting Chen},
  colorlinks=true,
  linkcolor={blue},
  filecolor={Maroon},
  citecolor={Blue},
  urlcolor={Blue},
  pdfcreator={LaTeX via pandoc}}

\title{Toward Embedding-Based Psychometrics: Structural Modeling of
Assessment-Item Semantics With Contextual Scores}
\author{Jinsong Chen and Shi-Ting Chen \tabularnewline Faculty of Education, The University of Hong Kong \tabularnewline Correspondence: jinsong.chen@live.com}
\date{}
\begin{document}
\maketitle
\begin{abstract}
Contextual scores represent assessment items through their similarities
to reference words in an external corpus. We examine the semantic
structure of scores for 40 TIMSS mathematics scored units using a
partially specified two-step factor procedure. A search across factor
counts identifies a persistent seven-group structure under the featured
construction. Subsequent comparisons consistently favor a general
dimension alongside group associations, although individual group
memberships remain sensitive to some specification choices. Item
examples distinguish recurring, cross-domain, sensitive, and imposed
associations. Simpler and unrestricted references clarify the
contribution and limits of the anchored representation: it improves on a
single factor but does not achieve the lowest working Bayesian
information criterion (BIC). A separate response benchmark compares
three initial Q constructions and their Hull-PVAF revisions under
higher-order and saturated attribute distributions. Among these
diagnostic models, BIC favors the official content framework and the
Akaike information criterion (AIC) favors its direct four-factor
augmentation, but a matched unidimensional two-parameter logistic model
has lower AIC and BIC than all twelve conditions. These findings support
a conditional semantic representation while limiting direct diagnostic
interpretation. We discuss learned text-assisted response calibration as
a prospective application requiring a larger calibrated item bank and
independent evaluation.

\end{abstract}

\section{Introduction}\label{sec-intro}

Assessment items carry structured information before anyone answers
them. Their wording, mathematical objects, and task contexts relate them
to other items and to a wider domain of mathematical language.
Psychometric modeling can make those relations explicit: it can impose
partial substantive structure, estimate remaining associations, separate
general from group variation, and investigate which features persist
when the model changes. The scientific question is what these semantic
representations establish about items, and what additional evidence
connects them to behavioral measurement.

Text embeddings provide one input to this investigation. Distributional
semantics represents meaning through context
(\citeproc{ref-harris1954distributional}{Harris, 1954}), and modern
encoders map item text to contextual vectors
(\citeproc{ref-ethayarajh2019contextual}{Ethayarajh, 2019};
\citeproc{ref-reimers2019sentence}{Reimers \& Gurevych, 2019}). Related
applications include item-difficulty prediction, scoring, generation,
and knowledge-component recovery
(\citeproc{ref-attali2022interactive}{{Attali et al.}, 2022};
\citeproc{ref-benedetto2021transformers}{Benedetto et al., 2021};
\citeproc{ref-shen2021classifying}{Shen et al., 2021};
\citeproc{ref-sung2019pretraining}{Sung et al., 2019}). Here,
\emph{contextual scores} measure the similarity between each item and
reference words whose contextual representations are averaged over an
external corpus (\citeproc{ref-chen2025documents}{J. Chen, 2025}). A
factor model of this score matrix describes corpus-conditioned semantic
relations among assessment items. Its factors are not examinee
abilities.

The contribution concerns the structure and interpretation of this
analyzed object. Partial framework specification designates some
item--factor associations while leaving others selectable. Persistence
across wider factor models addresses a different question from choosing
the best-fitting count. Comparing correlated factors with a
general-plus-groups representation examines how common variation is
organized. Signed loadings and model-conditional inclusion support then
describe a graded semantic item--factor map. Together, these operations
permit a more explicit investigation than treating every embedding
coordinate or similarity as interchangeable psychometric evidence.

The underlying ideas have distinct origins. Contextual-score work
proposes psychometric modeling of text
(\citeproc{ref-chen2025documents}{J. Chen, 2025}). Partially exploratory
factor analysis (PEFA), partially confirmatory factor analysis (PCFA),
and the two-step methodology supply the estimators and structural
comparison rules (\citeproc{ref-chen2021partially}{J. Chen et al.,
2021}; \citeproc{ref-chen2022generalized}{J. Chen, 2022},
\citeproc{ref-chen2023fully}{2023},
\citeproc{ref-chenjin2026stable}{2026}; \citeproc{ref-chenjin2026fit}{J.
Chen \& Jin, 2026a}; \citeproc{ref-jin2025regularized}{Jin \& Chen,
2025}). The present application adapts and evaluates these components
for assessment-item semantics, reversing the original document/word
orientation and examining the connection to student responses. It does
not introduce those inherited estimators as new methods.

Item content can indicate the knowledge and operations a successful
response requires. This motivates testing whether a structured semantic
representation helps describe response patterns. The connection is
empirical: items with similar wording or mathematical contexts may still
require different solution processes. A diagnostic Q-matrix gives binary
item--attribute requirements (\citeproc{ref-rupp2010diagnostic}{Rupp et
al., 2010}; \citeproc{ref-tatsuoka1983rule}{Tatsuoka, 1983}). Such
requirements have a stronger interpretation than selected semantic
associations. Expert construction and response-driven estimation or
revision supply different kinds of evidence
(\citeproc{ref-chiu2009cluster}{Chiu et al., 2009};
\citeproc{ref-delatorre2008empirically}{{de la Torre}, 2008};
\citeproc{ref-delatorrechiu2016general}{{de la Torre \& Chiu}, 2016};
\citeproc{ref-liu2012data}{Liu et al., 2012}). We therefore examine
binary projections as a consequential use of the semantic
representation, while retaining the graded map as an output in its own
right. A direct framework-guided PCFA supplies a simpler four-domain
comparator to the seven-group projection. Student responses evaluate
both these initial text-based constructions and the official content
framework, and subsequently revise separate copies through Hull-PVAF
(\citeproc{ref-najera2021hull}{{Nájera et al.}, 2021}).

This sequential design also differs from embedding-informed cognitive
diagnosis that uses text to form a prior and learns Q entries jointly
with responses (\citeproc{ref-liu2026scalable}{Liu et al., 2026}).
Contextual scores are the primary data in the text analyses here.
Responses do not construct their scores, factors, markers, or initial
binary maps; responses do participate in the explicitly labeled Hull
revisions. The separation makes it possible to investigate the
behavioral relevance of a response-free representation and the changes
introduced by empirical revision.

Three questions organize the application:

\begin{enumerate}
\def\labelenumi{\arabic{enumi}.}
\tightlist
\item
  \textbf{Semantic structure:} What continuous structure can be
  represented in contextual scores, and which aspects persist across
  factor counts and construction definitions?
\item
  \textbf{Structured item representation:} How do substantive anchors,
  general-versus-group representation, and selection support organize
  item--factor associations, and which associations recur across
  definitions?
\item
  \textbf{Relationship to behavioral measurement:} How do the initial Q
  constructions and their response-informed revisions compare in
  response fit and complexity under higher-order and saturated attribute
  distributions, and how do those models compare with a matched
  unidimensional reference?
\end{enumerate}

These questions support an empirical investigation of embedding-based
psychometrics. Favorable response comparisons can supply validity
evidence for a specified use of the text-based structure; they do not
establish every selected association as a cognitive requirement.
Unfavorable comparisons delimit that use without deciding whether every
possible graded representation is unhelpful.

The practical distinction matters for a larger item bank. A semantic
model can describe items before administration, whereas a response model
estimates how students perform on them. Using the former to inform the
latter requires an empirically learned connection. We return in the
Discussion to contextual scores and continuous item loadings as possible
predictors or prior information for response parameters. That
prospective use motivates further work; the present application
evaluates semantic structure and the narrower direct-Q use.

\section{Data and Contextual Scores}\label{sec-data}

\subsection{TIMSS Items and Student
Responses}\label{timss-items-and-student-responses}

The application uses 40 text-only scored units from the released U.S.
TIMSS 2011 grade-eight mathematics assessment
(\citeproc{ref-mullis2012timss}{Mullis et al., 2012}). Two shared stems
contain separately scored parts, so the units represent 36 base items.
The official content framework distinguishes number, algebra, geometry,
and data and chance; the cognitive framework distinguishes knowing,
applying, and reasoning. These frameworks describe different aspects of
an item and provide the alternative substantive starting points examined
in the structural search. Table \ref{tbl-crosswalk} in Appendix A fixes
item numbering and reports format, official classifications, response
coverage, and descriptive unidimensional two-parameter logistic (2PL)
estimates.

Student responses follow a booklet design, so each student answers only
a subset of the focal items. Table \ref{tbl-boundaries} distinguishes
the contextual-score matrix from the response matrices and their
analysis-specific samples. Missing response cells retain their original
booklet pattern. The descriptive 2PL calibration and residual checks use
students with at least one observed focal response; the common Q
benchmark requires at least two. Student responses do not enter score
construction, factor estimation, or any initial Q.

\begin{longtable}[]{@{}
  >{\raggedright\arraybackslash}p{(\linewidth - 4\tabcolsep) * \real{0.3333}}
  >{\raggedright\arraybackslash}p{(\linewidth - 4\tabcolsep) * \real{0.3333}}
  >{\raggedright\arraybackslash}p{(\linewidth - 4\tabcolsep) * \real{0.3333}}@{}}
\caption{Construction and evaluation
matrices.}\label{tbl-boundaries}\tabularnewline
\toprule\noalign{}
\begin{minipage}[b]{\linewidth}\raggedright
Feature
\end{minipage} & \begin{minipage}[b]{\linewidth}\raggedright
Contextual-score matrix
\end{minipage} & \begin{minipage}[b]{\linewidth}\raggedright
Student-response matrix
\end{minipage} \\
\midrule\noalign{}
\endfirsthead
\toprule\noalign{}
\begin{minipage}[b]{\linewidth}\raggedright
Feature
\end{minipage} & \begin{minipage}[b]{\linewidth}\raggedright
Contextual-score matrix
\end{minipage} & \begin{minipage}[b]{\linewidth}\raggedright
Student-response matrix
\end{minipage} \\
\midrule\noalign{}
\endhead
\bottomrule\noalign{}
\endlastfoot
Row unit & Frequent-word semantic profile across 40 items & Student
responses across administered items \\
Origin & Computed conditional on encoder, corpus, template, and
vocabulary & Sampled behavior under the TIMSS booklet design \\
Row relation & Word profiles share corpus and encoder geometry; no
i.i.d. sampling claim is made & Students are observational units under a
complex booklet design; item coverage is sparse \\
Primary outputs & Semantic loadings, posterior inclusion support, and
descriptive item-factor associations & Item parameters and residual
dependence \\
Role here & Constructs and perturbs the text-side representation &
Evaluates limited behavioral bridges only \\
\end{longtable}

\subsection{Contextual-Score Construction and
Interpretation}\label{contextual-score-construction-and-interpretation}

Following J. Chen (\citeproc{ref-chen2025documents}{2025}) with the
original document--word roles reversed, frequent corpus words index the
rows and TIMSS items index the measured columns. This orientation lets
the analysis investigate relations among items through a common
reference vocabulary. Words play an algebraic role analogous to cases in
a factor-analysis matrix; they are computed, corpus-conditioned
profiles, rather than sampled persons. The featured reference corpus is
MathQA (\citeproc{ref-amini2019mathqa}{Amini et al., 2019}). Documents
were tokenized and lemmatized, and stopwords and numeric-only tokens
were removed. The vocabulary rule and realized matrix dimensions are
reported in Appendix A.

Qwen3-Embedding-8B maps both items and word occurrences into a common
vector space (\citeproc{ref-zhang2025qwen3embedding}{Zhang et al.,
2025}). Each clean item document contains the problem, response options
where present, and key, but no generated rationale or official framework
label. Item embeddings use last-token pooling. Word occurrences use a
different operation: final-layer hidden states for subword tokens
overlapping the target occurrence are averaged; occurrences are then
averaged within each document, followed by an equal-weight average
across containing documents. This produces the mean contextual embedding
\(\mathbf{w}_i\) of word \(i\). Equal weighting prevents a document with
repeated occurrences from dominating the reference word. If
\(\mathbf{h}_j\) denotes the embedding of item \(j\), Equation
\ref{eq-score} defines the featured cosine-similarity contextual score:

\begin{equation}\protect\phantomsection\label{eq-score}{
y^{(\mathrm{cos})}_{ij}=\frac{\mathbf{w}_i^{\mathsf T}\mathbf{h}_j}
{\lVert\mathbf{w}_i\rVert\lVert\mathbf{h}_j\rVert}.
}\end{equation}

The score is defined even when word \(i\) does not appear in item \(j\);
larger values indicate greater semantic alignment. The unnormalized
score \(y^{(\mathrm{dot})}_{ij}=\mathbf{w}_i^{\mathsf T}\mathbf{h}_j\)
is retained as a metric-form sensitivity. It preserves vector magnitude
and is not a second featured construction or a new Step-1 sweep.

Thus \(\mathbf{Y}\) has 1,864 word rows and 40 item columns and is
standardized by column before factor analysis. Conditional on the
encoder, corpus, document template, and vocabulary rule, its entries are
computed rather than sampled person responses. The implications of that
distinction and the observed row-profile concentration are considered
below.

The construction can also be expressed geometrically. Let \(W\) and
\(H\) contain unit-normalized word and item embeddings, respectively,
and let \(C_n=I_n-\mathbf{1}\mathbf{1}^{\mathsf T}/n\) be the centering
matrix, where \(n\) is the number of word rows (1,864 in the featured
construction) and plays the role of the sample size. Then

\begin{equation}\protect\phantomsection\label{eq-geometry}{
Y=WH^{\mathsf T},\qquad
S_Y=\frac{1}{n-1}Y^{\mathsf T}C_nY
=H\left(\frac{1}{n-1}W^{\mathsf T}C_nW\right)H^{\mathsf T}.
}\end{equation}

The corpus and retained vocabulary therefore define the geometry through
which item embeddings are compared. Column standardization subsequently
rescales \(S_Y\) to the correlation matrix used in the factor analyses.
This identity explains why a corpus change can change the modeled item
relations; it does not identify the empirical general factor or turn
word rows into independent respondents. Recomputing the scores from the
saved embedding inputs reproduced both matrices under the original
single-precision matrix-product recipe.

The word-frequency cutoff is an application construction rule. The
stricter cutoffs examine dependence on the less frequent part of the
retained vocabulary; they do not establish a universally optimal
threshold. The MathQA-plus-GSM8K construction
(\citeproc{ref-cobbe2021gsm8k}{Cobbe et al., 2021}) is a nested corpus
sensitivity because it contains the featured MathQA corpus. A change of
corpus can therefore change the semantic question represented by the
matrix, as well as its numerical estimates.

The two data objects require different interpretations. A
contextual-score matrix can be modeled using the algebra of factor
analysis without inheriting the sampling interpretation of a
student-response matrix. Its factors summarize shared item semantics
relative to the reference vocabulary. Its uniquenesses describe
remaining semantic variation after the retained factors.

The diagnostics in Appendix A, Table \ref{tbl-score-diagnostics}, assess
item-column distributions and the concentration of word profiles. The
contextual score distributions have some asymmetry and excess kurtosis.
These summaries make the working Gaussian approximation inspectable.
Comparing them with binary response marginals does not amount to testing
a shared sampling model.

The full contextual matrix shows substantial raw elevation and residual
row-profile redundancy after standardization: participation-ratio rank
rises from 2.86 to 13.20 (Table \ref{tbl-score-diagnostics}). The
matched comparison is equally important: word and response profiles have
broadly overlapping concentration summaries once the same booklet
columns are used. Thus row-profile correlation does not by itself
diagnose stochastic dependence. The evidential boundary instead follows
from construction: words share an encoder and corpus and are not an
independent random sample, whereas students are sampled behavioral
units. The participation-ratio rank remains a spectral concentration
index, not an effective sample size. Conventional likelihood and
absolute-fit indices are therefore treated descriptively on the
contextual-score side.

There is no conventional person-sampling noise floor for the contextual
scores; residual variance represents semantic structure not captured by
the retained factors rather than examinee-level measurement error. No
score-matrix standard error or significance test is interpreted as if
the rows were independent. Absolute-fit indices and the Bayesian
information criterion (BIC) are descriptive under a working
independent-row factor likelihood. Step 1 therefore emphasizes count
paths, direct persistence, and perturbations. In Step 2, the evidential
result is the repeated direction of four fixed within-definition
contrasts outside the tie band, not their numerical magnitude as a Bayes
factor (\citeproc{ref-chenjin2026fit}{J. Chen \& Jin, 2026a}).

The analytical procedure also distinguishes within-fit anchoring,
post-fit alignment, and diagnostic-Q identification. Their roles are
explained in the methods; Appendix B records the associated design and
numerical checks.

\section{Analytical Approach}\label{sec-method}

The procedure separates two analytical decisions that a single
exploratory fit would conflate. Step 1 uses a PEFA sweep to ask whether
a factor count yields a stable structure that persists under direct
comparison with wider solutions. Step 2 holds the selected count fixed
and asks whether the same group definition is better represented as
oblique factors or as a bifactor structure with a general dimension. The
resulting signed loadings and inclusion support define the graded
semantic item--factor representation. Section \ref{sec-response} then
examines what happens when selected group associations are used as
binary entries in a response model.

\begin{figure}

\centering{

\includegraphics[width=1\linewidth,height=\textheight,keepaspectratio]{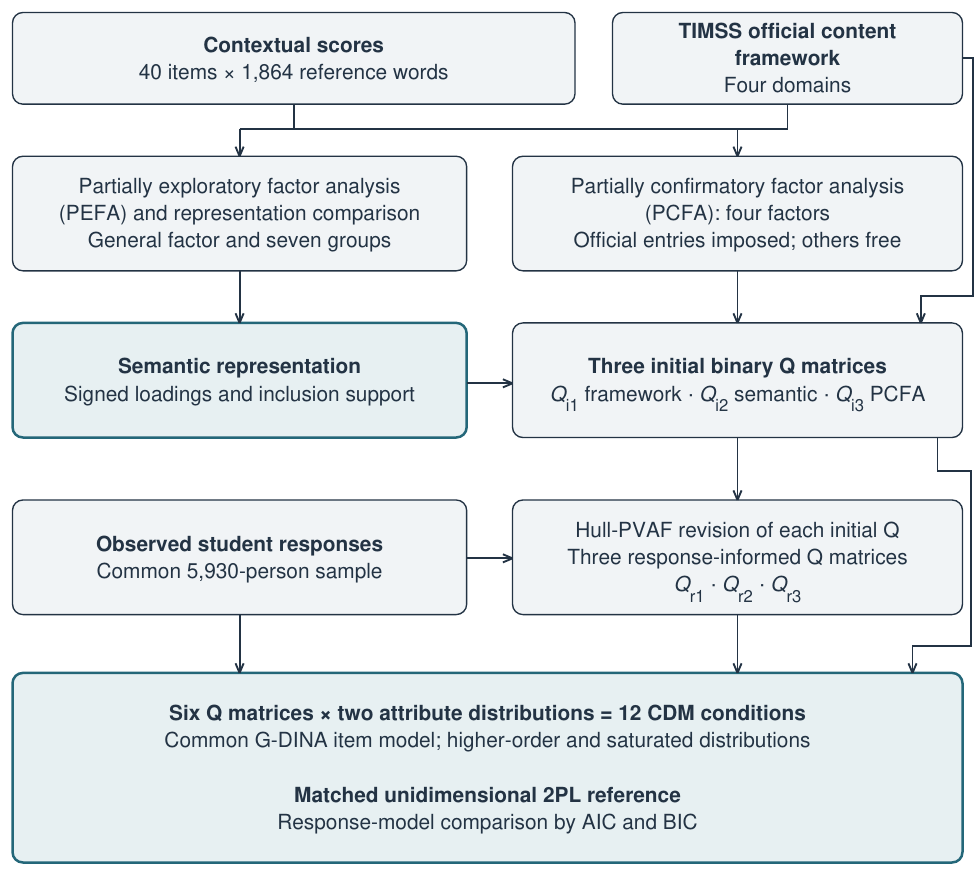}

}

\caption{\label{fig-pipeline}Analytical flow. Contextual scores support
the graded semantic map as the primary structural output. Three
response-free binary constructions enter the response benchmark.
Hull-PVAF uses student responses to revise separate Q copies; all six
mappings are evaluated under higher-order and saturated attribute
distributions, alongside a matched 2PL reference. Diagnostic eligibility
is assessed separately from the existence of the graded representation.}

\end{figure}%

\subsection{Identifying Persistent
Structure}\label{identifying-persistent-structure}

For standardized \(\mathbf{Y}\), the model is
\(\mathbf{Y}=\mathbf{F}\boldsymbol\Lambda^{\mathsf T}+\mathbf{E}\) with
\(\boldsymbol\Sigma=\boldsymbol\Lambda\boldsymbol\Phi
\boldsymbol\Lambda^{\mathsf T}+\boldsymbol\Psi\). A loading-design
matrix marks each loading as specified present (1), fixed zero (0), or
unspecified (-1). Unspecified loadings receive a continuous
spike-and-slab prior. Under the mean-field variational Bayes (VB)
approximation, the posterior inclusion probability (PIP) is the
posterior probability that a loading belongs to the slab
(active-loading) component (\citeproc{ref-george1997approaches}{George
\& McCulloch, 1997}; \citeproc{ref-guoqiang2022variational}{Lin, 2022};
\citeproc{ref-rovckova2014emvs}{Ročková \& George, 2014}). The fitted
software is \texttt{vbpm} (\citeproc{ref-chen2026vbpm}{J. Chen \& Jin,
2026b}). Included loadings have a \(N(0,1)\) prior; unspecified loadings
use slab variance 1 and the spike-variance path in Table
\ref{tbl-estimation-settings}. The inclusion-rate prior is
\(\mathrm{Beta}(1,1)\) and residual precisions use a Gamma prior with
shape 1 and rate .1. Oblique covariance updates use the implementation
settings \(\xi_\Phi=2K+1\) and \(\Lambda_\Phi=I_K\), followed by
standardization to a correlation matrix; the full settings are
deposited. We reserve \emph{Q-matrix} for a later binary diagnostic
candidate; the ternary object used to fit the factor model is called the
loading design.

The featured anchor-only (AO) design fixes selected official-content
anchors as present and leaves every other loading selectable. The
complete Step-1 grid crosses content versus cognitive backbones, MathQA
versus the consolidated corpus, AO versus an anchor-zero (AZ)
sensitivity, and nominal depths two, four, and six. AZ additionally
fixes cross-factor zeros; it remains in the complete Step-1 census
because those zeros can displace broad elevation into the unspecified
columns and alter their relation to the backbone. The featured
specification is MathQA, content, AO, depth four. Content-backbone cells
fit candidate counts \(K=4,\ldots,14\); cognitive-backbone cells fit
\(K=3,\ldots,14\), giving 276 candidate fits across the 24
specifications. No column is interpreted substantively during the sweep.

Here \(K\) is the total number of Step-1 group columns, including four
content or three cognitive backbone columns and excluding the general
column introduced only in a bifactor Step-2 fit. Nominal depth is the
target number of author-selected anchors per backbone factor, capped by
the number of eligible items in that category. The anchor ladder is
development-informed rather than prospectively specified. Items were
ranked by author judgment of category prototypicality; extension beyond
depth two incorporated preliminary stability evidence, and item M032295
was excluded as an algebra anchor because preliminary loadings raised a
label-alignment concern. The cell and ladder are therefore neither
independently selected nor independently validated, and no outcome-blind
validation of that ladder is claimed.

PEFA is used here because it preserves the official content-framework
backbone while allowing the number and composition of added columns to
change across the sweep. A fully unrestricted extraction would not
preserve that reference structure.

Count evidence and structural persistence are evaluated separately.
Increases in the evidence lower bound (ELBO) and decreases in BIC are
normalized by each path's largest positive gain. After the last maximum,
the first transition \(K\rightarrow K+1\) with gain below 20\% returns
\(K\); a single crossing suffices. Without a crossing, no count is
returned. The finite readings form the count-evidence set
\(\mathcal{K}_{20\%}\), whose subscript records the 20\% gain threshold.
For comparison of \(K\) with \(K+s\), backbone columns are paired by
position. Each added column independently matches the added column with
the smallest squared loading distance after sign alignment. Reusing a
matched column is a collision and fails the comparison. Persistence is
the minimum absolute congruence over matched columns. Following J. Chen
(\citeproc{ref-chenjin2026stable}{2026}), a persistence profile
\((r_c,m)\) pairs a cutoff \(r_c\), which this minimum congruence must
reach, with a horizon \(m\): \(K\) persists when its comparison with
each \(K+s\), \(s=1,\ldots,m\), is collision-free and reaches \(r_c\).
Each comparison is direct rather than a chain of adjacent matches. The
three nonnested profiles are \((r_c,m)=(.85,1)\), \((.80,2)\), and
\((.70,3)\), labeled .85/one, .80/two, and .70/three in the tables; they
trade a stricter cutoff over a shorter horizon against a laxer cutoff
over a longer one.

Persistence is assessed for \(K=4,\ldots,11\) in content specifications
and \(K=3,\ldots,11\) in cognitive specifications; higher fitted counts
supply the required wider comparisons. Within each profile, these
candidate counts are examined from largest to smallest. \(K_p\) is the
highest persistent count only after every larger candidate has been
shown to fail the persistence rule; an unavailable comparison at a
larger candidate leaves the conclusion unresolved. A persistent \(K_p\)
in \(\mathcal{K}_{20\%}\) is L1. A persistent \(K_p\) outside
\(\mathcal{K}_{20\%}\), with at least one finite count reading, is L2.
If every candidate fails the persistence rule, the result is L3.
Unresolved persistence, or a persistent \(K_p\) without a finite count
reading, remains unclassified. L1 and L2 therefore identify persistent
structures, with count uncertainty in L2. Agreement of the three
profiles on \(K_p\) and layer is a separate robustness result
(Table~\ref{tbl-layers}).

\Needspace{240pt}

\begin{longtable}[]{@{}
  >{\raggedright\arraybackslash}p{(\linewidth - 6\tabcolsep) * \real{0.2500}}
  >{\raggedright\arraybackslash}p{(\linewidth - 6\tabcolsep) * \real{0.2500}}
  >{\raggedright\arraybackslash}p{(\linewidth - 6\tabcolsep) * \real{0.2500}}
  >{\raggedright\arraybackslash}p{(\linewidth - 6\tabcolsep) * \real{0.2500}}@{}}
\caption{Logical layers for one persistence profile. Cross-profile
agreement is reported separately.}\label{tbl-layers}\tabularnewline
\toprule\noalign{}
\begin{minipage}[b]{\linewidth}\raggedright
Layer
\end{minipage} & \begin{minipage}[b]{\linewidth}\raggedright
Count evidence
\end{minipage} & \begin{minipage}[b]{\linewidth}\raggedright
Structural evidence
\end{minipage} & \begin{minipage}[b]{\linewidth}\raggedright
Reporting outcome
\end{minipage} \\
\midrule\noalign{}
\endfirsthead
\toprule\noalign{}
\begin{minipage}[b]{\linewidth}\raggedright
Layer
\end{minipage} & \begin{minipage}[b]{\linewidth}\raggedright
Count evidence
\end{minipage} & \begin{minipage}[b]{\linewidth}\raggedright
Structural evidence
\end{minipage} & \begin{minipage}[b]{\linewidth}\raggedright
Reporting outcome
\end{minipage} \\
\midrule\noalign{}
\endhead
\bottomrule\noalign{}
\endlastfoot
L1 & \(K_p\in \mathcal{K}_{20\%}\) & Persistent & Persistent structure
with count support \\
L2 & \(K_p\notin \mathcal{K}_{20\%}\) and \(\mathcal{K}_{20\%}\) is
nonempty & Persistent & Persistent structure with count uncertainty \\
L3 & Count paths may be usable or unusable & All candidate counts fail
the persistence rule & No persistent structure identified \\
Unclassified & Any & Persistence unresolved, or persistent \(K_p\) with
no finite count reading & Report the evidence limitation; do not relabel
it L3 \\
\end{longtable}

The count-to-persistence distance
\(g_{CP}=\min_{k\in \mathcal{K}_{20\%}}|K_p-k|\) is descriptive: it does
not determine the layer or overturn an L2 finding. In this application,
\(g_{CP}\ge3\) is flagged as beyond the count discrepancy examined in
the methodological framework. Such results remain L2 but are excluded
from Step 2 in this application. Same-count depth sensitivities and
uniform marker-depth eligibility are applied only after the Step-1
readings are known; their outcomes are reported in
Section~\ref{sec-step1}.

\subsection{Comparing Factor
Representations}\label{comparing-factor-representations}

Conditional on an eligible L1/L2 proposal, Step 2 uses
\(K_{\mathrm{step2}}=K_p\). The application evaluates four fixed
definitions at the retained count: backbone depths two and four crossed
with two marker rules. The overlap-permitted rule independently assigns
each added group the same number of markers as the inherited backbone
depth, choosing the largest absolute Step-1 loadings. Inherited anchors
remain eligible, and any item may be reused across added groups. The
globally disjoint rule solves a single joint assignment across the three
added groups, uses the same depth, and forbids reuse; it is not three
independent factorwise choices. Each definition is fitted as (a)
correlated group factors and (b) an anchored bifactor model with a
general factor and orthogonal group factors. The representation contrast
compares BIC within each definition, with \(|\Delta\mathrm{BIC}|\le2\)
treated as a practical tie. BIC does not select \(K\) or backbone depth.
A separate, disclosed within-depth comparison of the two depth-four
bifactor fits selects the marker rule used for the adopted map.

Step 2 is not independent confirmation: it reuses the same computed
score matrix, and its group markers are inherited from Step 1. The
featured specification and downstream family were not preregistered. A
within-depth marker-rule choice based on bifactor BIC is therefore
labeled data-adaptive and application-specific; the realized
eligibility, representation verdict, and adoption are reported in
Section~\ref{sec-step2}.

The correlated representation allows relationships among group factors,
with factor variances standardized to one. The bifactor representation
instead assigns common variation to a general factor and retains
mutually orthogonal group factors; its factor covariance matrix is the
identity. The general column is present for every item. These are
different decompositions of the same item covariance, not an assertion
that one set of student abilities is nested within another. A
specified-present loading is estimated under its loading prior; the
design entry 1 imposes presence, not a numerical loading equal to one.

Two reference models help interpret the comparison: a single general
factor and an unrestricted model with the same total factor count as the
adopted bifactor. A correlated PCFA guided directly by the four official
content domains provides an additional reference. For that PCFA,
official memberships are present and all other loading cells are
selectable, with no specified zeros, general factor, or count sweep.
This separates the effect of elaborating the framework through the full
procedure from the simpler use of contextual scores to augment its
existing domains.

For each item-by-group cell, the signed loading records direction and
magnitude, whereas the PIP records model-conditional variational support
that a free loading belongs to the slab (active-loading) component. We
report these layers together as a \emph{graded semantic item--factor
map}. The graded representation is an uncertainty-aware semantic
item--factor map for inspection and potential future expert drafting.
The term does not imply a response-calibrated cognitive requirement
probability: PIP is conditional on the contextual-score matrix, loading
design, variational prior, and fitted representation. A
specified-present marker has no estimated inclusion probability because
its presence was imposed. This object also differs from Bayesian
uncertain-Q models, whose posterior probabilities concern latent binary
Q entries learned with responses (\citeproc{ref-chen2018bayesianq}{Y.
Chen et al., 2018}; \citeproc{ref-decarlo2012uncertainty}{DeCarlo,
2012}).

PIP is unsigned, so thresholding a positive or negative selected loading
cannot turn a continuous compensatory association into a positive
conjunctive requirement.

The posterior-thresholded binary item-factor map is a descriptive
projection of the graded semantic map. A fixed-present group cell equals
1; an unspecified group cell equals 1 when its raw, unrenormalized PIP
is at least .50. A value of 1 therefore means that an item is associated
with a factor under the statistical rule. It does not mean that experts
have confirmed the factor as a required diagnostic attribute. The rule
is cellwise, permits multiple memberships, and never replaces them by an
argmax partition.

Cross-definition mapping stability is evaluated only after the seven
group columns are aligned. Each aligned item-by-group cell is classified
as always free, always specified-present, or mixed across the four
designs. PIP correlations and PIP-.50 agreement and Jaccard indices use
the always-free cells only. For such a cell,
\(S_{{\rm free},jk}=\sum_s I({\rm PIP}_{jks}\ge .50)\) counts support
across the four fitted definitions: zero and four denote nonselection
and selection in every definition, whereas one through three denote
threshold variation. This count is a descriptive sensitivity summary,
not a vote or a rule for constructing a consensus mapping.

\subsection{Estimation and Evaluation}\label{estimation-and-evaluation}

For loading design \(D\) and raw free-cell PIP \(\pi\), define
\(A_{jk}=\mathbb{1}\{D_{jk}=1\ \mathrm{or}\ (D_{jk}=-1,\pi_{jk}\ge .50)\}\).
The hard-selected covariance used by the reported criterion is

\begin{equation}\protect\phantomsection\label{eq-hard-covariance}{
\Sigma_H=(\Lambda\odot A)\Phi(\Lambda\odot A)^{\mathsf T}+\Psi.
}\end{equation}

This is a variational plug-in covariance; the selected model is not
refitted by maximum likelihood. Both the Gaussian working likelihood
scale and the BIC penalty use \(n-1\), so
\(\mathrm{BIC}=-2\ell_H+p\log(n-1)\). Thus the inclusion threshold
enters text-side fit assessment as well as the later binary projection.
The corresponding Akaike information criterion (AIC) uses the penalty
\(2p\). Raw free-cell PIPs are not renormalized. Parameter counts and
the local covariance-Jacobian check are reported with the structural
comparisons; the exact accounting and numerical tolerance appear in
Appendix B. A full local rank concerns the selected covariance
parameterization at a fitted point. It does not establish global
rotational identification, unthresholded posterior identification, or
diagnostic-Q identification.

Factor composition is described from signed loadings before
thresholding. For signed reporting, the general column is oriented to
have positive mean loading; each adopted group column is oriented so
that the sum of its signed loadings over cells selected by the
fixed-present/PIP-.50 rule is positive, with the largest-absolute
loading positive as a tie fallback. This reporting convention does not
change fit, PIPs, the binary projection, or sign-invariant
correspondence; group signs remain relative polarities rather than
invariant cognitive directions. For cross-definition correspondence,
only \(G\) is fixed. All seven group columns are matched independently
by minimum sign-invariant squared error, after which target distinctness
is checked. A reused target column would be recorded as a collision and
would make the comparison inadmissible; no collision is repaired by a
global assignment. Group design positions are not fixed across
definitions because the marker rules differ and the same positional
label need not retain the same empirical column. This
similarity-optimized rule was adopted after an observed mismatch showed
that positional matching was inadequate; it is a post-estimation
descriptive alignment, not prespecified or independent stability
evidence.

Three operations must remain separate. Specified markers provide
operational anchoring within a fit, but do not prove a unique global
loading matrix for the exact design. Alignment is applied only after
estimation to make cross-definition reporting possible. Diagnostic-Q
identification concerns the later binary object and requires evidence
beyond either operation.

The adopted seven-group bifactor object is refitted under
document-frequency thresholds 35 and 50, dot-product similarity,
alternative loading-prior paths, random keyword halves, and twenty 80\%
keyword subsamples. These are fixed-count, fixed-design continuous
sensitivity analyses. They do not repeat Step 1 or establish persistence
recovery under those new inputs.

The aligned PEFA sweep and regularized PCFA representations are
implemented within the variational estimator. This provides an
executable common framework, not proof that VB is necessary or superior.
Mean-field approximation can understate uncertainty; PIPs and their
downstream decisions remain conditional on the likelihood, prior, and
approximation. The five Step-2 calls with different seed labels use
deterministic initialization and produce repeated evaluations, not
diverse random starts. The response-model optimizer checks use separate
randomized starts. Appendix A reports recorded candidate timings and a
larger-bank component cost grid, including separate R-memory
measurements. These provide partial computational evidence; they do not
measure the complete two-step procedure or establish a general
complexity law.

Variational approximation also matters along the analytical chain.
Attenuated loadings can change congruence and the ranking used to select
markers. Understated uncertainty can sharpen PIPs and therefore affect
both the selected covariance and binary projection. Applying the same
estimator to several models does not ensure that these effects cancel.
Prior-path perturbations inspect conditional sensitivity; they do not
estimate variational bias or replace a comparison with a more complete
posterior calculation. The current evidence therefore supports
inspection of fitted patterns, with uncertainty interpreted under the
stated approximation.

Response analyses use their own respondent likelihood and calibration
controls. They enter only after the initial semantic representations
have been constructed. The Q benchmark is described in Section
\ref{sec-q-benchmark}; additional response checks and their distinct
targets are documented in Appendix D. The analysis flow is summarized in
Figure \ref{fig-pipeline}, and Appendix A supplies the stage-specific
settings. In each case, a fitted model, a repeatable pattern, and a
useful interpretation are separate evidential questions.

\section{Semantic Structure of the Items}\label{sec-structure-results}

\subsection{Persistence Across Factor Counts}\label{sec-step1}

The featured construction yields a seven-group structure that persists
under each of the declared profiles. Count evidence is less uniform: the
working-BIC path supports the persistent count, whereas the ELBO path
favors a wider model. The result therefore combines structural
persistence with qualified count support. Appendix B, Table
\ref{tbl-trajectory}, reports the complete count path and adjacent
comparisons. Absolute-fit trajectories are deposited for all candidate
fits; their conventional respondent-sample cutoffs are not used to
select a count from computed word profiles.

Table \ref{tbl-transitions} shows the direct comparisons underlying the
persistence decision. The seven-column solution remains recognizable in
wider solutions even where an adjacent transition is weaker. In
particular, following the eight-factor solution into the nine-factor
solution gives a different impression from comparing the original
seven-column structure directly with the same wider solution. This
example explains why persistence cannot be inferred by chaining adjacent
similarities. It asks whether the original structure is retained, rather
than whether every intermediate enlargement is equally stable.

\Needspace{175pt}

\begin{longtable}[]{@{}rrrr@{}}
\caption{Direct persistence
horizons.}\label{tbl-transitions}\tabularnewline
\toprule\noalign{}
Factor count \(K\) & \(\phi(K,K+1)\) & \(\phi(K,K+2)\) &
\(\phi(K,K+3)\) \\
\midrule\noalign{}
\endfirsthead
\toprule\noalign{}
Factor count \(K\) & \(\phi(K,K+1)\) & \(\phi(K,K+2)\) &
\(\phi(K,K+3)\) \\
\midrule\noalign{}
\endhead
\bottomrule\noalign{}
\endlastfoot
4 & .986 & .695 & .647 \\
5 & .451 & .342 & .322 \\
6 & .928 & .855 & .878 \\
7 & .900 & .926 & .924 \\
8 & .742 & .604 & .385 \\
9 & .589 & .325 & .384 \\
10 & .355\(^{\dagger}\) & .183\(^{\dagger}\) & .072 \\
11 & .293 & .260\(^{\dagger}\) & .014\(^{\dagger}\) \\
\end{longtable}

\begin{minipage}{\linewidth}
\footnotesize\textit{Note.} Depth four. Every cell compares the solution at $K$ directly with the stated wider solution; $\dagger$ marks a matching collision.
\end{minipage}

Table~\ref{tbl-persistence} applies the profiles to all three depths. At
depth four, all profiles independently return \(K_p=7\)/L1: the same
seven-factor solution persists across one, two, and three direct
horizons while belonging to \(\mathcal{K}_{20\%}\). Depth two identifies
an L2 structure but changes count in the deep profile. Depth six
identifies a persistent structure only under the shallow profile and
cannot support a uniform depth-six marker design because of its
category-size caps, with realized backbone depths of 6/6/4/6. Thus depth
four is primary; depth-two \(K=7\) is a same-count Step-2 sensitivity,
not a competing main specification.

\begin{longtable}[]{@{}
  >{\raggedleft\arraybackslash}p{(\linewidth - 10\tabcolsep) * \real{0.1429}}
  >{\centering\arraybackslash}p{(\linewidth - 10\tabcolsep) * \real{0.1786}}
  >{\centering\arraybackslash}p{(\linewidth - 10\tabcolsep) * \real{0.1786}}
  >{\centering\arraybackslash}p{(\linewidth - 10\tabcolsep) * \real{0.1786}}
  >{\centering\arraybackslash}p{(\linewidth - 10\tabcolsep) * \real{0.1786}}
  >{\raggedleft\arraybackslash}p{(\linewidth - 10\tabcolsep) * \real{0.1429}}@{}}
\caption{Depth-by-profile
persistence.}\label{tbl-persistence}\tabularnewline
\toprule\noalign{}
\begin{minipage}[b]{\linewidth}\raggedleft
Depth
\end{minipage} & \begin{minipage}[b]{\linewidth}\centering
\(\mathcal{K}_{20\%}\)
\end{minipage} & \begin{minipage}[b]{\linewidth}\centering
Profile
\end{minipage} & \begin{minipage}[b]{\linewidth}\centering
Persistent counts
\end{minipage} & \begin{minipage}[b]{\linewidth}\centering
\(K_p\)/layer
\end{minipage} & \begin{minipage}[b]{\linewidth}\raggedleft
\(g_{CP}\)
\end{minipage} \\
\midrule\noalign{}
\endfirsthead
\toprule\noalign{}
\begin{minipage}[b]{\linewidth}\raggedleft
Depth
\end{minipage} & \begin{minipage}[b]{\linewidth}\centering
\(\mathcal{K}_{20\%}\)
\end{minipage} & \begin{minipage}[b]{\linewidth}\centering
Profile
\end{minipage} & \begin{minipage}[b]{\linewidth}\centering
Persistent counts
\end{minipage} & \begin{minipage}[b]{\linewidth}\centering
\(K_p\)/layer
\end{minipage} & \begin{minipage}[b]{\linewidth}\raggedleft
\(g_{CP}\)
\end{minipage} \\
\midrule\noalign{}
\endhead
\bottomrule\noalign{}
\endlastfoot
2 & 8, 9 & .85/one & 5, 7 & 7/L2 & 1 \\
2 & 8, 9 & .80/two & 5, 7 & 7/L2 & 1 \\
2 & 8, 9 & .70/three & 5 & 5/L2 & 3 \\
4 & 7, 9 & .85/one & 4, 6, 7 & 7/L1 & 0 \\
4 & 7, 9 & .80/two & 6, 7 & 7/L1 & 0 \\
4 & 7, 9 & .70/three & 6, 7 & 7/L1 & 0 \\
6 & 5, 9 & .85/one & 5 & 5/L1 & 0 \\
6 & 5, 9 & .80/two & none & none/L3 & -- \\
6 & 5, 9 & .70/three & none & none/L3 & -- \\
\end{longtable}

\begin{minipage}{\linewidth}
\footnotesize\textit{Note.} L1 and L2 identify persistent structures; L3 is resolved nonpersistence. Depth-two $K=5$ meets the $g_{CP}\ge3$ flag and is therefore excluded from Step 2; depth six is excluded for the reasons stated in text.
\end{minipage}

The complete construction census in Appendix B, Table
\ref{tbl-step1-census}, shows that persistence is not confined to the
featured cell. Other framework and corpus combinations also yield
persistent solutions, often with uncertainty about the supported count.
The distinctive featured result is agreement on the same seven-group,
count-supported solution across all profiles. All candidate fits
converged, but convergence alone does not explain this contrast between
construction choices.

Anchor-only specifications more often preserve a stable structure than
designs that impose cross-factor zeros. Those zeros can relocate broad
semantic elevation into added columns, changing their relation to the
framework columns. This is informative about the consequences of
specification, rather than evidence that every other construction fails.
The census uses cosine scores throughout; the word-frequency and
dot-product checks in Section \ref{sec-robust} are fixed-design refits
and answer a different question. The profile readings share data and
models, so they are not independent replications.

\subsection{General and Group Dimensions}\label{sec-step2}

Every examined group definition favors a bifactor representation over
its correlated-factor counterpart under the working BIC rule (Table
\ref{tbl-step2}). The direction recurs across changes in anchor depth
and marker assignment. The common result is that variation shared
broadly across items should be represented explicitly, leaving group
factors to describe more specific associations. The size of each
contrast remains definition dependent.

\Needspace{300pt}

\begin{longtable}[]{@{}
  >{\raggedright\arraybackslash}p{(\linewidth - 10\tabcolsep) * \real{0.1304}}
  >{\centering\arraybackslash}p{(\linewidth - 10\tabcolsep) * \real{0.2174}}
  >{\raggedleft\arraybackslash}p{(\linewidth - 10\tabcolsep) * \real{0.1739}}
  >{\raggedleft\arraybackslash}p{(\linewidth - 10\tabcolsep) * \real{0.1739}}
  >{\raggedleft\arraybackslash}p{(\linewidth - 10\tabcolsep) * \real{0.1739}}
  >{\raggedright\arraybackslash}p{(\linewidth - 10\tabcolsep) * \real{0.1304}}@{}}
\caption{Step-2 representation fits.}\label{tbl-step2}\tabularnewline
\toprule\noalign{}
\begin{minipage}[b]{\linewidth}\raggedright
Definition
\end{minipage} & \begin{minipage}[b]{\linewidth}\centering
Effective \(p\) (O/B)
\end{minipage} & \begin{minipage}[b]{\linewidth}\raggedleft
Oblique BIC
\end{minipage} & \begin{minipage}[b]{\linewidth}\raggedleft
Bifactor BIC
\end{minipage} & \begin{minipage}[b]{\linewidth}\raggedleft
\(\Delta\)BIC
\end{minipage} & \begin{minipage}[b]{\linewidth}\raggedright
Verdict
\end{minipage} \\
\midrule\noalign{}
\endfirsthead
\toprule\noalign{}
\begin{minipage}[b]{\linewidth}\raggedright
Definition
\end{minipage} & \begin{minipage}[b]{\linewidth}\centering
Effective \(p\) (O/B)
\end{minipage} & \begin{minipage}[b]{\linewidth}\raggedleft
Oblique BIC
\end{minipage} & \begin{minipage}[b]{\linewidth}\raggedleft
Bifactor BIC
\end{minipage} & \begin{minipage}[b]{\linewidth}\raggedleft
\(\Delta\)BIC
\end{minipage} & \begin{minipage}[b]{\linewidth}\raggedright
Verdict
\end{minipage} \\
\midrule\noalign{}
\endhead
\bottomrule\noalign{}
\endlastfoot
Depth 4, globally disjoint & 174/173 & 92,910.9 & 89,432.5 & 3,478.4 &
Bifactor \\
Depth 4, overlap permitted & 175/173 & 92,646.2 & 90,182.8 & 2,463.3 &
Bifactor \\
Depth 2, overlap permitted & 174/171 & 91,800.9 & 89,702.4 & 2,098.5 &
Bifactor \\
Depth 2, globally disjoint & 174/180 & 91,800.8 & 90,094.9 & 1,705.9 &
Bifactor \\
\end{longtable}

\begin{minipage}{\linewidth}
\footnotesize\textit{Note.} Seven group factors. O/B = oblique/bifactor; effective $p$ is the rank-adjusted BIC parameter count. $\Delta$BIC is oblique minus bifactor and is compared only within a definition; it does not select depth or factor count. Differences are computed before rounding.
\end{minipage}

The primary depth and adopted marker rule have different justifications.
The depth-four construction was development informed and provides the
common count-supported persistent structure described above. Within that
depth, the disjoint-marker bifactor has lower BIC and supplies the
adopted semantic map. Table \ref{tbl-step2} also shows why this is an
application-specific choice: the preferred marker rule reverses at the
shallower depth. The alternative definitions remain part of the
analysis, rather than being discarded after selection. Their comparisons
do not establish a generally optimal anchor depth or marker rule.

Two additional references use the same standardized score matrix,
estimator, diagonal residual specification, and working likelihood
convention: a single factor with all item loadings included, and an
unrestricted eight-factor model with all 320 loadings included under the
estimator's Gaussian loading prior. The latter has orthogonal factors
and no framework anchoring or selected zeros. Varimax rotation is used
only to describe its loadings; it does not change its fitted covariance
or criteria. The direct four-factor PCFA used for the third Q is also
shown for context in Table~\ref{tbl-r51-text-references}.

\begin{longtable}[]{@{}
  >{\raggedright\arraybackslash}p{(\linewidth - 10\tabcolsep) * \real{0.2200}}
  >{\raggedleft\arraybackslash}p{(\linewidth - 10\tabcolsep) * \real{0.1200}}
  >{\raggedleft\arraybackslash}p{(\linewidth - 10\tabcolsep) * \real{0.1700}}
  >{\raggedleft\arraybackslash}p{(\linewidth - 10\tabcolsep) * \real{0.1700}}
  >{\raggedleft\arraybackslash}p{(\linewidth - 10\tabcolsep) * \real{0.1700}}
  >{\raggedleft\arraybackslash}p{(\linewidth - 10\tabcolsep) * \real{0.1500}}@{}}
\caption{Text-model references on the common contextual-score
matrix.}\label{tbl-r51-text-references}\tabularnewline
\toprule\noalign{}
\begin{minipage}[b]{\linewidth}\raggedright
Model
\end{minipage} & \begin{minipage}[b]{\linewidth}\raggedleft
Factors
\end{minipage} & \begin{minipage}[b]{\linewidth}\raggedleft
Nominal/rank \(p\)
\end{minipage} & \begin{minipage}[b]{\linewidth}\raggedleft
Working AIC
\end{minipage} & \begin{minipage}[b]{\linewidth}\raggedleft
Working BIC
\end{minipage} & \begin{minipage}[b]{\linewidth}\raggedleft
Hard SRMR
\end{minipage} \\
\midrule\noalign{}
\endfirsthead
\toprule\noalign{}
\begin{minipage}[b]{\linewidth}\raggedright
Model
\end{minipage} & \begin{minipage}[b]{\linewidth}\raggedleft
Factors
\end{minipage} & \begin{minipage}[b]{\linewidth}\raggedleft
Nominal/rank \(p\)
\end{minipage} & \begin{minipage}[b]{\linewidth}\raggedleft
Working AIC
\end{minipage} & \begin{minipage}[b]{\linewidth}\raggedleft
Working BIC
\end{minipage} & \begin{minipage}[b]{\linewidth}\raggedleft
Hard SRMR
\end{minipage} \\
\midrule\noalign{}
\endhead
\bottomrule\noalign{}
\endlastfoot
One factor & 1 & 80/80 & 125,201.8 & 125,644.2 & .0712 \\
Framework PCFA & 4 & 133/133 & 103,758.2 & 104,493.7 & .0622 \\
Retained oblique & 7 & 174/174 & 91,948.7 & 92,910.9 & .0676 \\
Retained bifactor & 8 & 173/173 & 88,475.8 & 89,432.5 & .0278 \\
Unrestricted eight & 8 & 360/332 & 83,570.3 & 85,406.2 & .0574 \\
\end{longtable}

The reference models sharpen the interpretation of the two-step result
(Table \ref{tbl-r51-text-references}). A single factor leaves
substantial structure unexplained. The retained bifactor improves
considerably on that reference, but the unrestricted model has lower
working BIC than the anchored solution. Conversely, the anchored
bifactor has smaller standardized covariance residuals. The likelihood
criterion and unweighted residual summary emphasize different aspects of
approximation, so these rankings can disagree.

This comparison places a clear limit on the methodological claim. The
anchored procedure supplies a designated substantive orientation and
explicit inclusion decisions; it does not optimize every measure of
covariance fit. The unrestricted reference accounts for rotational
freedom in its rank-adjusted complexity. Recognizable content and shared
context can emerge under either orientation. The item interpretations
below therefore concern what the fitted representation makes
inspectable, rather than claiming that its orientation is uniquely
correct.

\subsection{Interpreting Item--Factor Associations}\label{sec-mapping}

The general factor is positive across the items and accounts for the
largest share of squared loading mass (Table \ref{tbl-composition}).
This quantity describes the loading matrix; it is not the share of
observed variance explained and is not a measure of students' general
mathematics ability. Group columns retain more specific relations after
common semantic variation is represented. Several have a leading
official content domain, while others mix domains or concentrate on
shared item context.

\begin{table}[H]
\centering
\begingroup\singlespacing
\setlength{\tabcolsep}{2.8pt}
\renewcommand{\arraystretch}{1.02}
\fontsize{9.0}{10.5}\selectfont
\begin{threeparttable}
\caption{Factor composition and effects.}
\label{tbl-composition}
\begin{tabularx}{\textwidth}{@{}lrrr>{\raggedright\arraybackslash}p{3.5cm}>{\raggedright\arraybackslash}X@{}}
\toprule
Factor & \shortstack{SSL\\share} & \shortstack{Relative\\effect\\$D/D_G$} & Mapped & Three largest loadings & Composition or stem concentration \\
\midrule
G  & 75.8\% & 1.000 & 40 & 23 (+.929), 6 (+.924), 10 (+.919) & Positive global elevation \\
S1 & 6.4\%  & .175 & 19 & 29 (+.607), 27 (+.569), 19 (+.442) & Algebra 11; number 7; geometry 1 \\
S2 & 3.9\%  & .152 & 13 & 35 (+.720), 40 (+.640), 15 (+.327) & Data 6; algebra 4; number 3 \\
S3 & 3.6\%  & .185 & 10 & 31 (+.717), 34 (+.661), 32 (+.270) & Geometry 4; data 3; number 3 \\
S4 & 1.3\%  & .056 & 11 & 11 (+.338), 6 (+.271), 5 (+.188) & Number 8; algebra 3 \\
S5 & 3.0\%  & .086 & 18 & 25 (+.472), 24 (+.361), 14 (+.332) & Broad/sign mixed; M042198 share = 39.9\% \\
S6 & 3.8\%  & .198 & 10 & 38 (+.615), 37 (+.603), 36 (+.580) & M042169 share = 81.0\% \\
S7 & 2.1\%  & .088 & 12 & 33 (+.404), 13 (+.283), 32 (+.274) & Mixed; number 6, algebra 3 \\
\bottomrule
\end{tabularx}
\begin{tablenotes}[flushleft]\fontsize{9.0}{10.5}\selectfont
\item Note. Depth-four globally disjoint bifactor fit. SSL share is each column's squared-loading sum divided by the total over all eight columns. $D$ is the mean squared loading over cells selected by design or raw PIP $\ge .50$; $D_G$ is the general-column mean over all 40 items, so G = 1. Mapped is the corresponding item count. Item numbers refer to Table \ref{tbl-crosswalk}. Composition statements are hypotheses for expert review, not factor names.
\end{tablenotes}
\end{threeparttable}
\endgroup
\end{table}

The distinction between the two prominent shared-stem patterns is
substantive. For the staff-count items, the associated group is
concentrated on the common setting and closely related statistical
tasks. For the sequence items, the corresponding group also includes
substantial associations outside that stem. Treating both as
interchangeable testlet factors would erase this difference. The group
labels therefore remain descriptive coordinates. Content and stem
concentrations are hypotheses that can be inspected against item
wording, with response dependence assessed separately.

Each item--factor association can be read through two complementary
quantities: its signed loading describes direction and magnitude, and
its free-cell PIP describes variational support for inclusion under the
model. Imposed markers are distinguished from estimated support. The
complete displays in Appendix C retain both quantities, allowing readers
to inspect strong loadings, weak but selected associations, and cells
whose apparent certainty follows from the design.

The adopted fit separates many included and excluded cells clearly, yet
such separation within one fit does not ensure stability across
definitions. The examples in Table \ref{tbl-r51-examples} illustrate the
meanings and limitations of these different kinds of association. They
are descriptions of item semantics, not response-calibrated
probabilities that a cognitive attribute is required.

Four items illustrate what a selected association can mean. These
examples were chosen by declared descriptive rules: the largest absolute
adopted loading among always-free associations selected across all four
definitions; the largest positive PCFA addition outside an item's
official content domain; the adopted selected always-free association
nearest PIP .50 among cells selected in one to three definitions; and
the largest absolute S6 loading within the M042169 stem. Selection
criteria and full numerical records accompany the archive. The examples
explain estimated patterns; they are not independent expert validation.

\begin{longtable}[]{@{}
  >{\raggedright\arraybackslash}p{(\linewidth - 4\tabcolsep) * \real{0.3333}}
  >{\raggedright\arraybackslash}p{(\linewidth - 4\tabcolsep) * \real{0.3333}}
  >{\raggedright\arraybackslash}p{(\linewidth - 4\tabcolsep) * \real{0.3333}}@{}}
\caption{Substantive readings of recurring, cross-domain, sensitive, and
imposed associations.}\label{tbl-r51-examples}\tabularnewline
\toprule\noalign{}
\begin{minipage}[b]{\linewidth}\raggedright
Item and task
\end{minipage} & \begin{minipage}[b]{\linewidth}\raggedright
Association
\end{minipage} & \begin{minipage}[b]{\linewidth}\raggedright
Interpretation
\end{minipage} \\
\midrule\noalign{}
\endfirsthead
\toprule\noalign{}
\begin{minipage}[b]{\linewidth}\raggedright
Item and task
\end{minipage} & \begin{minipage}[b]{\linewidth}\raggedright
Association
\end{minipage} & \begin{minipage}[b]{\linewidth}\raggedright
Interpretation
\end{minipage} \\
\midrule\noalign{}
\endhead
\bottomrule\noalign{}
\endlastfoot
M042245: identify the equation fitting two coordinate pairs & Adopted S1
loading .607, PIP 1.000; free and selected in all four definitions & The
algebra-centered group captures a recognizably algebraic task. \\
M042186: extend a subtraction pattern to a negative subtrahend &
Official number domain; PCFA algebra addition .923, PIP 1.000 &
Contextual scores connect an arithmetic item with algebraic pattern
extension. The initial PCFA Q retains its official number membership. \\
M032047: express the sum of three consecutive integers when the middle
is \(2n\) & Adopted S6 loading .100, PIP .662; free and selected in two
of four definitions & A formally algebraic expression can have a weak,
definition-sensitive secondary association. \\
M042169C: compare effects on the mean and median after increasing the
largest staff count & Adopted S6 loading .615; imposed marker & Shared
context and statistical content both contribute plausible
interpretations. This imposed cell has no estimated PIP. \\
\end{longtable}

The coordinate-pair item provides the most straightforward reading. Its
task asks the student to recognize an algebraic rule, and its recurring
association with the algebra-centered group is estimated freely across
definitions. Agreement between task content and the fitted pattern makes
the association interpretable. It does not by itself establish that the
factor is a necessary or sufficient response attribute, but it shows how
the model can expose a recognizable relation without imposing that
particular cell.

The subtraction-pattern item illustrates a different contribution. Its
official classification emphasizes number, whereas its continuation task
also involves recognizing a pattern. The direct four-domain PCFA retains
the official membership and estimates an additional algebra association.
This is a useful example of semantic augmentation: the model makes a
cross-domain relation explicit rather than forcing the item into a
single exclusive category. Whether that relation improves a response
model is a subsequent empirical question.

The consecutive-integers item is intentionally retained as a less
decisive case. Its secondary association is weak and changes selection
across definitions. Reading the PIP alone would obscure the small signed
loading; reading the binary projection alone would additionally hide the
variation across definitions. The graded representation lets these
qualifications remain attached to the item. The example demonstrates why
every selected cell should not receive an equally strong substantive
interpretation.

Finally, the staff-count item is a specified marker. Its association
helps interpret the shared-context group, but its inclusion cannot count
as a discovery of that relation. The marker and the freely estimated
algebra association thus illustrate different roles of substantive
information: one orients a factor, whereas the other provides an
estimated relation to be inspected. Keeping both visible prevents a
persuasive item narrative from substituting for evidence about which
cells the model actually learned.

\subsection{Stability and Sensitivity}\label{sec-robust}

The recurring result is a general-plus-groups representation, with
stronger agreement in the general profile than in individual group
associations. Table \ref{tbl-stability} brings the main comparisons
together. Centered profile correlations are more informative here than
near-unit uncentered congruence: when all general loadings are positive,
a large common level can make two profiles look very similar even if
their relative item pattern changes.

Group comparisons require alignment because alternative markers can
exchange or reorient columns. Every examined pair admits a
collision-free match, but alignment does not make the group patterns
invariant. The full pairwise matrix is deposited with the analysis
archive. The support comparison uses only cells free under every
definition, so imposed markers do not inflate the count of recurring
estimated associations. Some associations recur, some remain unselected,
and a substantial subset changes its inclusion decision.

\begin{table}[H]
\centering
\begingroup\singlespacing
\fontsize{10.5}{12.2}\selectfont
\setlength{\tabcolsep}{4pt}
\renewcommand{\arraystretch}{1.08}
\caption{Stability and sensitivity of the continuous representation.}
\label{tbl-stability}
\begin{tabularx}{\textwidth}{@{}p{.22\textwidth}p{.38\textwidth}X@{}}
\toprule
Comparison & Numerical result & Interpretation\\
\midrule
General profiles across definitions & Centered $r=.9141$--.9978; $\phi\ge .9987$ & Broad common dimension with some relative-profile variation\\
Matched group profiles & Minimum $\phi=.5986$--.8797 across definition pairs & Less uniform group patterns\\
237 always-free cells & 33 selected throughout; 145 never selected; 59 change status & Estimated-cell recurrence, excluding imposed markers\\
Frequency cutoff 35 & 1,330 words; minimum group $\phi=.9947$ & Close agreement conditional on the adopted design\\
Frequency cutoff 50 & 1,080 words; minimum group $\phi=.9919$ & Close agreement conditional on the adopted design\\
Dot-product scores & Minimum group $\phi=.7655$; centered G $r=.8631$ & Metric choice changes some associations\\
Nonannealed prior & Minimum group $\phi=.0861$ positionally, .6259 after matching & Column exchange explains part of the discrepancy\\
Random keyword halves & Minima: group $\phi=.9851$, G $\phi=.9999$, centered G $r=.9952$ & Close agreement under keyword splitting\\
Twenty 80\% keyword subsamples & Minima: group $\phi=.7862$, G $\phi=.9992$, centered G $r=.9403$; the other 19 group minima range from .9894 to .9987 & Strong matching in most refits, with one weaker group result\\
\bottomrule
\end{tabularx}
\par\vspace{3pt}\parbox{\textwidth}{\footnotesize Note. $\phi$ is uncentered congruence; $r$ is the centered correlation of general-loading profiles. Vocabulary, metric, prior, and resampling refits hold the factor count and loading design constant.}
\endgroup
\end{table}

Random keyword halves and most subsamples preserve close agreement,
although one subsample changes the group pattern more substantially. The
stricter vocabulary refits also preserve the loading pattern closely.
This is evidence that the adopted representation does not depend heavily
on the less frequent retained words, conditional on keeping its count
and design fixed. It is a different claim from recovering the same
persistent count after repeating the complete search. The latter claim
has not been tested by these refits. The full construction census and
the fixed-design sensitivities should therefore be read together, with
their different targets kept explicit.

Normalization and prior specification affect some groups more strongly.
The dot-product score preserves embedding magnitude as well as
directional alignment, so it changes the analyzed information. The prior
perturbation also exchanges two columns. Matching removes that
positional artifact but leaves meaningful pattern differences. The
corrected comparison supports neither an apparent collapse based on
unmatched positions nor complete invariance after alignment.

These findings reconcile persistence with construction dependence.
Corpus and framework choices define the semantic relations under
investigation; sensitivity checks show how those relations behave under
a specified perturbation. In addition, the featured specification and
anchor ladder were developed using information from this application.
Their favorable behavior is not independent confirmation. The recurring
broad organization is conditional on this bank and construction; its
sensitive associations remain part of the result. Evaluating transfer
requires construction choices made before examining a new bank.

\section{Response Evaluation}\label{sec-response}

The response analyses examine direct use of semantic associations in a
binary Q, after the text-based representation has been constructed. The
benchmark is a test of that specified application. It neither defines
every possible use of contextual scores nor establishes that a
different, untested use would succeed.

\subsection{Q-Matrix Constructions and Comparison
Design}\label{sec-q-benchmark}

Three initial Q matrices operationalize different uses of item
information. We write \(Q_{\mathrm{i}1}\), \(Q_{\mathrm{i}2}\), and
\(Q_{\mathrm{i}3}\) for these initial Qs and \(Q_{\mathrm{r}1}\),
\(Q_{\mathrm{r}2}\), and \(Q_{\mathrm{r}3}\) for their Hull-PVAF
revisions: the subscripts i and r mark the initial and revised stages,
and the numbers 1 to 3 the origin. The official content framework
\(Q_{\mathrm{i}1}\) assigns each item to one of algebra, data and
chance, geometry, and number, giving 40 memberships. The semantic group
projection \(Q_{\mathrm{i}2}\) uses the adopted seven group columns,
excluding G, and the stated PIP rule, giving 93 memberships. The
framework-guided PCFA \(Q_{\mathrm{i}3}\) comes directly from a
four-factor correlated PCFA of the featured contextual scores. Its
design sets all 40 official-domain cells to 1 and every other cell to
\(-1\); there are no specified-zero entries, general factor, or
factor-count sweep. Specified cells remain included and free cells enter
the binary projection at raw PIP \(\ge .50\).

The PCFA can augment official memberships but cannot remove one during
the initial text-only construction. Table \ref{tbl-q-inventory}
summarizes the three initial maps and the realized revision sizes.
Appendix D distinguishes the imposed PCFA cells from selected additions,
including additions with negative signed loadings. A selected semantic
cross-loading does not itself establish a cognitive requirement. The
seven-group map has different column meanings and is not forced into an
entrywise alignment with the official four domains.

Each initial Q receives one common Hull-PVAF revision using saturated
G-DINA calibration (\citeproc{ref-najera2021hull}{{Nájera et al.},
2021}). The algorithm balances item fit and parsimony within the
supplied attribute space; it does not discover the correct number of
attributes. We use test-attribute iteration and retain only a Q
unchanged across successive revisions. The same revised Q is then fitted
under both attribute distributions. Thus, the planned comparison
contains three origins, two Q stages, and two attribute distributions:
12 fitted conditions. A loop, empty attribute, exhausted iteration
limit, or failed calibration leaves the corresponding revised conditions
unavailable rather than triggering an ad hoc repair.

\begin{longtable}[]{@{}
  >{\raggedright\arraybackslash}p{(\linewidth - 8\tabcolsep) * \real{0.3000}}
  >{\raggedleft\arraybackslash}p{(\linewidth - 8\tabcolsep) * \real{0.1200}}
  >{\raggedleft\arraybackslash}p{(\linewidth - 8\tabcolsep) * \real{0.1700}}
  >{\raggedleft\arraybackslash}p{(\linewidth - 8\tabcolsep) * \real{0.2400}}
  >{\raggedleft\arraybackslash}p{(\linewidth - 8\tabcolsep) * \real{0.1700}}@{}}
\caption{Initial Q constructions and empirical revision
inventory.}\label{tbl-q-inventory}\tabularnewline
\toprule\noalign{}
\begin{minipage}[b]{\linewidth}\raggedright
Construction
\end{minipage} & \begin{minipage}[b]{\linewidth}\raggedleft
Attributes
\end{minipage} & \begin{minipage}[b]{\linewidth}\raggedleft
Initial memberships
\end{minipage} & \begin{minipage}[b]{\linewidth}\raggedleft
Hull additions/removals
\end{minipage} & \begin{minipage}[b]{\linewidth}\raggedleft
Revised memberships
\end{minipage} \\
\midrule\noalign{}
\endfirsthead
\toprule\noalign{}
\begin{minipage}[b]{\linewidth}\raggedright
Construction
\end{minipage} & \begin{minipage}[b]{\linewidth}\raggedleft
Attributes
\end{minipage} & \begin{minipage}[b]{\linewidth}\raggedleft
Initial memberships
\end{minipage} & \begin{minipage}[b]{\linewidth}\raggedleft
Hull additions/removals
\end{minipage} & \begin{minipage}[b]{\linewidth}\raggedleft
Revised memberships
\end{minipage} \\
\midrule\noalign{}
\endhead
\bottomrule\noalign{}
\endlastfoot
Official content framework (\(Q_{\mathrm{i}1}\), \(Q_{\mathrm{r}1}\)) &
4 & 40 & 3 / 1 & 42 \\
Semantic group projection (\(Q_{\mathrm{i}2}\), \(Q_{\mathrm{r}2}\)) & 7
& 93 & 0 / 2 & 91 \\
Framework-guided PCFA (\(Q_{\mathrm{i}3}\), \(Q_{\mathrm{r}3}\)) & 4 &
87 & 2 / 3 & 86 \\
\end{longtable}

All twelve CDM fits use the identical 5,930-by-40 matrix without
sampling weights, retaining its booklet missingness and the G-DINA
item-response kernel. Higher-order (HO) fits use a standard-normal
higher-order variable with freely estimated positive attribute slopes;
saturated fits estimate unrestricted population class proportions. If
item \(j\) has \(K_j\) memberships and the Q has \(K\) columns, the
parameter counts are \(\sum_j 2^{K_j}+2K\) for HO and
\(\sum_j 2^{K_j}+2^K-1\) for saturated fits. Item probabilities are
unconstrained by monotonicity in both arms.

The calibrations use common GDINA and Hull-PVAF controls, reported in
Appendix A and the analysis archive. Multiple requested starts and one
permitted longer run address optimizer convergence. The Hull
implementation applies the same calibration controls to internal fits
and catches optional final absolute-fit errors under the sparse booklet
design. Its Q-selection and iteration logic are unchanged from the
recorded package source. The initial and revised maps therefore differ
through the declared empirical revision procedure rather than through ad
hoc repairs.

BIC is the primary relative criterion; AIC, log likelihood, and
parameter counts are reported alongside it. Computational eligibility
requires convergence and finite likelihoods, probabilities, and
item-parameter standard errors. Higher-order structural standard errors
were not estimated. Boundary cells and local probability decreases are
reported, rather than used to discard otherwise eligible fits.
Probability residuals compare observed item and coadministered-pair
proportions with expectations integrated over the fitted population
attribute distribution. The available-pair summaries describe fit to
observed margins; they are not a complete-matrix global-fit test.

Initial-Q comparisons can contribute response-based validity evidence
for the specified text-to-structure-to-Q procedure: responses were
absent from those constructions. Comparing Qs with different attribute
spaces evaluates each joint Q/model specification, so a ranking cannot
isolate the contextual scores alone. Hull-revised comparisons evaluate
response-informed refinement, and their conventional AIC/BIC count the
fitted model parameters while omitting the preceding adaptive search
over Q entries. These full-sample comparisons do not measure held-out
prediction, recovery of a true Q, or individual mastery accuracy.

A unidimensional 2PL fitted to the same observed response matrix
provides a reference outside the CDM family. Both packages report
observed-data marginal likelihoods, integrating over their respective
continuous or discrete latent variables. Those different latent
representations do not inherently prevent information-criterion
comparison: they model the same observed responses. The comparison
requires compatible likelihood definitions and constants, observations
and missing-response treatment, likelihood maximization, and parameter
counts. We treat it as an exploratory cross-package comparison of
nonnested models, rather than a likelihood-ratio test. Boundary
estimates, identification concerns, and uncounted Hull search cost limit
the formal interpretation of the criteria.

\subsection{Comparative Results and Their
Implications}\label{comparative-results-and-their-implications}

The matched unidimensional 2PL has lower AIC and BIC than all twelve CDM
conditions, including both stages of the official-framework Q
(\(Q_{\mathrm{i}1}\) and \(Q_{\mathrm{r}1}\)) (Table
\ref{tbl-r51-benchmark}). Several text-derived CDMs have higher
in-sample likelihoods, but these gains do not offset their additional
parameters under either criterion. Thus the 2PL has the better relative
fit--complexity balance; the ranking does not establish absolute
adequacy, true unidimensionality, or individual diagnostic accuracy.

All planned CDM conditions converge with finite likelihoods,
probabilities, and item-parameter standard errors, and each Hull branch
reaches a stable revised Q. The saturated \(Q_{\mathrm{i}2}\) and
\(Q_{\mathrm{r}2}\) fits require the longer iteration allowance. The
small revisions leave the main differences in attribute count and
membership density intact (Table \ref{tbl-q-inventory}).

\begin{table}

\caption{\label{tbl-r51-benchmark}Response-model comparisons on the common sample.}

\centering{

\centering
\begingroup\singlespacing
\fontsize{11}{13}\selectfont
\setlength{\tabcolsep}{4pt}
\renewcommand{\arraystretch}{1.05}

\begin{tabular}{@{}llrrrr@{}}
\toprule
\multicolumn{6}{@{}l}{\textit{Panel A. CDM comparisons}}\\[2pt]
Q/model & Dist. & $p$ & log $L$ & AIC & BIC \\
\midrule
$Q_{\mathrm{i}1}$ & Sat. & 95 & -30,896.98 & 61,983.96 & 62,619.30 \\
$Q_{\mathrm{i}1}$ & HO & 88 & -30,933.93 & 62,043.87 & 62,632.39 \\
$Q_{\mathrm{r}1}$ & Sat. & 99 & -30,873.91 & 61,945.82 & \textbf{62,607.91} \\
$Q_{\mathrm{r}1}$ & HO & 92 & -30,908.08 & 62,000.16 & 62,615.43 \\
$Q_{\mathrm{i}2}$ & Sat. & 417 & -30,486.56 & 61,807.13 & 64,595.93 \\
$Q_{\mathrm{i}2}$ & HO & 304 & -30,589.01 & 61,786.02 & 63,819.11 \\
$Q_{\mathrm{r}2}$ & Sat. & 411 & -30,496.64 & 61,815.27 & 64,563.95 \\
$Q_{\mathrm{r}2}$ & HO & 298 & -30,593.88 & 61,783.76 & 63,776.72 \\
$Q_{\mathrm{i}3}$ & Sat. & 223 & -30,614.67 & 61,675.34 & 63,166.71 \\
$Q_{\mathrm{i}3}$ & HO & 216 & -30,632.00 & 61,695.99 & 63,140.55 \\
$Q_{\mathrm{r}3}$ & Sat. & 217 & -30,585.29 & \textbf{61,604.58} & 63,055.83 \\
$Q_{\mathrm{r}3}$ & HO & 210 & -30,612.84 & 61,645.68 & 63,050.11 \\
\midrule
\multicolumn{6}{@{}l}{\textit{Panel B. Matched 2PL reference}}\\[2pt]
2PL & -- & 80 & -30,694.44 & 61,548.88 & 62,083.90 \\
\bottomrule
\end{tabular}
\par\vspace{3pt}\parbox{\textwidth}{\footnotesize Note. $N=5{,}930$ and 40 items throughout. Initial ($Q_{\mathrm{i}}$) and Hull-revised ($Q_{\mathrm{r}}$) Qs are numbered by origin as in Table \ref{tbl-q-inventory}. Sat. and HO denote saturated and higher-order attribute distributions; all Panel A models use the G-DINA item kernel. Bold values are minima within the twelve CDM conditions only. The 2PL has the lowest AIC and BIC overall, lower by 55.70 and 524.00 than the respective CDM minima. Criteria condition on the realized Q and omit Hull search cost. The cross-package comparison uses reported observed-data marginal likelihoods and is exploratory. CDM boundary and residual diagnostics appear in Appendix D.}
\endgroup

}

\end{table}%

Within the CDM family, BIC favors the official content framework, which
uses no contextual-score information, whereas AIC favors its direct PCFA
augmentation. The PCFA additions improve likelihood enough to satisfy
AIC's penalty, but not BIC's stronger penalty relative to the framework
Q. The semantic Qs (\(Q_{\mathrm{i}2}\), \(Q_{\mathrm{r}2}\)) have lower
AIC and higher BIC than the framework Qs (\(Q_{\mathrm{i}1}\),
\(Q_{\mathrm{r}1}\)) in all four matched comparisons, and higher values
of both criteria than the PCFA Qs (\(Q_{\mathrm{i}3}\),
\(Q_{\mathrm{r}3}\)) throughout. These distinctions qualify the
seven-group projection without making any CDM the overall preferred
model.

The ordering of Q origins under each criterion is unchanged between
higher-order and saturated attribute distributions. Within each origin,
BIC favors the higher-order distribution for the semantic and PCFA Qs
and the saturated distribution for the framework Qs, at both stages.
Hull lowers BIC in every matched comparison; the saturated
\(Q_{\mathrm{r}2}\) fit nevertheless has worse AIC than the saturated
\(Q_{\mathrm{i}2}\) fit.

Empirical edits also require substantive interpretation. The framework
revision moves M032132 from data and chance to algebra. The PCFA
revision removes the official algebra membership of M042077 and number
membership of M032626. Appendix Table \ref{tbl-r51-hull-changes} lists
all edits. They are response-driven fit decisions, not expert
confirmation of item requirements.

Boundary probabilities occur in both criterion-preferred CDMs and are
more widespread in the richer maps. Appendix D reports the CDM boundary
counts, higher-order slope limits, and residuals for coadministered
pairs. These diagnostics apply to the CDM fits; unreported 2PL
diagnostics are not zeros. Global absolute-fit calculations were
unavailable under the sparse booklet matrix. Convergence and
observed-pair residuals therefore support comparison of the fitted
conditions without establishing readiness for individual diagnosis.

The binary conversion removes magnitudes and signs, excludes the general
column, and treats selected associations as positive requirements.
Appendix C identifies the negative loadings retained by that conversion.
The diagnostic screen in Appendix D shows that the projection covers
every item but lacks exclusive indicators for S6 and S7; removing
shared-context columns creates other coverage problems. These
model-dependent flags do not prove nonidentification under general
G-DINA. Together with the response ranking, they limit direct diagnostic
use of the semantic map.

The additional response checks provide further qualifications.
Permutations favor the adopted item placement over arbitrary placements,
but that placement includes official-framework anchors as well as added
semantic markers. Continuous group geometry has only a weak association
with residual dependence, and grouped item-parameter prediction recovers
little linear signal in this small bank. Appendix D retains the
procedures, results, and parameter-precision context. These findings
neither isolate contextual-score construction as the source of the
unfavorable comparison nor demonstrate that a different use would
succeed.

\section{Discussion}\label{sec-discussion}

The application makes item-semantic information available for explicit
psychometric structural investigation. Its contribution is the
combination of a defined contextual-score object, partial substantive
specification, count/persistence distinctions, and an interpretable
separation between graded associations and response-model requirements.
This is a methodological application of existing contextual-score and
factor-modeling machinery, with empirical findings that support some
uses and constrain others.

The featured construction contains a persistent group structure, while
other construction cells also contain persistent solutions with
different count support. The direct comparisons illustrate why
persistence of a solution at a given count is distinct from similarity
at every adjacent enlargement. The representation contrasts then
consistently favor a general dimension alongside group associations.
These findings answer different questions: persistence concerns the
survival of a structure across counts, whereas the bifactor comparison
concerns how its common and more specific variation is organized.

The reference comparisons locate the contribution of anchoring in
substantive orientation and explicit selection: the unrestricted model
has better working BIC, whereas the anchored bifactor has smaller
covariance residuals. Neither result establishes improved attribute
recovery or reduced expert effort.

The graded map adds an item-level interpretation to those aggregate
findings. It distinguishes recurring freely estimated associations,
cross-domain augmentation, sensitive secondary associations, and imposed
markers. The examples show why these should not be read as
interchangeable forms of support. General profiles are more consistent
than individual group coordinates, and matching resolves some apparent
sensitivity without eliminating the remainder. The stability table and
full maps make these distinctions available for inspection without
requiring readers to infer them from a sequence of numerical
comparisons.

Construction sensitivity is scientifically consequential. Corpus and
vocabulary define the semantic geometry in Equation \ref{eq-geometry};
content and cognitive frameworks describe different aspects of an item.
It is plausible that lexical contextual scores align more readily with
content, but the present results do not identify the mechanism of that
difference. The featured specification and anchor ladder were
development informed, and the within-depth marker rule was selected
using the same score matrix. Complete census reporting exposes that
selection history. Fixed-design frequency and metric refits do not
establish that a new count sweep would recover the same structure. No
universal cutoff, independent replication, or chance-calibrated
guarantee against favorable selection is claimed.

The response benchmark tests the substantive premise that selected
semantic associations can serve as discrete mastery requirements. Among
CDMs, the official content framework is preferred by BIC and its direct
PCFA augmentation by AIC; the seven-group projection is less competitive
than the PCFA augmentation. The matched 2PL, however, has lower AIC and
BIC than every tested CDM. This includes the official-framework models,
so the unfavorable result is not specific to contextual-score
construction. It evaluates joint choices of Q, attribute definitions,
response model, and available item coverage; their separate
contributions remain unresolved. Information criteria do not measure
mastery accuracy or instructional benefit, and the present study
demonstrates neither. Weak residual and parameter-prediction
associations further constrain immediate behavioral interpretation.

The sampling and inferential boundaries also remain explicit. Computed
word rows share an encoder and corpus rather than forming an independent
respondent sample. Working likelihood comparisons and variational PIPs
remain conditional on that construction. The numerical covariance-rank
check concerns selected parameters at a fitted point; the RLCM witness
checks structural theorem conditions without certifying the observed
sparse-response model. The small, text-only item subset, missing
response patterns, boundary estimates, and incomplete encoder provenance
limit the claims that can be carried forward.

A prospective calibration application would learn a relationship between
text representations and parameters defined by a response model. A
sufficiently large calibrated subset could pair contextual scores,
embeddings, or signed loadings with response-derived Q memberships or
multidimensional item-response parameters. Predictions or informative
priors for new items could then be updated as responses accumulate. This
requires common identified response dimensions and uncertainty in both
estimated targets and predictions; Q memberships alone do not supply the
response parameters needed for cognitive diagnostic adaptive testing. A
small calibrated fraction must still contain enough varied items to
learn the relationship. More students answering the same few items
cannot replace that item-level information.

Published raw-embedding prediction results provide motivation for this
direction. In a larger mathematics bank, S.-T. Chen \& Chen
(\citeproc{ref-chenchen2026parameters}{2026}) report positive prediction
of 2PL difficulty and log-discrimination. That evidence concerns raw
embeddings and unidimensional response parameters, rather than
contextual-score loadings, Q recovery, or multidimensional adaptive
testing. Its repeated cross-validation assigns pooled items to folds
rather than holding out item families, so it does not establish
generalization to new item families or performance when near-duplicate
content is separated between training and evaluation. Future evaluation
should withhold whole items or families, vary the amount of calibration
data, and assess eventual measurement accuracy as well as parameter
prediction. The weak present prediction results leave success uncertain;
a larger bank enables a more informative test rather than guaranteeing
improvement.

Further research should also address a small number of distinct
questions. A construction specified before examining a new item bank can
test transfer across banks or corpora. Expert review can assess whether
semantic groups and cross-domain associations represent intended content
or shared wording. A response model that retains signed, graded semantic
information can test what is lost through binary projection; the current
experiment cannot attribute a shortfall to binarization alone.
Larger-bank validation and comparable MCMC estimation would address
transport and approximation uncertainty, respectively, rather than
serving as interchangeable remedies.

Embedding-based psychometrics can examine information available before
administration while making its relationship to response processes an
explicit empirical question. The present study supports a representation
conditional on this bank and construction, with persistent and sensitive
components, a direct framework-guided alternative, and response
comparisons of initial and empirically revised mappings. Its value lies
in making those distinctions inspectable and testable, with stronger
diagnostic interpretations requiring further evidence.

\section*{Data and Code Availability}\label{data-and-code-availability}
\addcontentsline{toc}{section}{Data and Code Availability}

The released TIMSS 2011 grade-eight mathematics items and response
database are available from the IEA TIMSS and PIRLS International Study
Center. An anonymized analysis archive is available at
\url{https://osf.io/7dv2f/overview?view_only=6af04daca10342ac846f85b1f1d53a59}.
The archive contains the featured contextual scores, aggregate result
records, initial and revised Q matrices, exact analysis settings,
verification and re-estimation scripts, and checksums. Its verifier
checks the deposited numerical results. Re-estimating the reference text
models requires the specified R packages; response re-estimation
additionally requires acquisition and preparation of the external TIMSS
data according to the documented instructions. Respondent-level
responses, identifiers, person scores, fitted CDM objects containing
responses, corpus documents, encoder weights, and released item text are
not redistributed.

The archive README distinguishes numerical verification from model
re-estimation and documents the upstream extraction limits and
external-data requirements.

\section*{References}\label{references}
\addcontentsline{toc}{section}{References}

\begingroup\singlespacing\fontsize{10.5}{12.0}\selectfont

\protect\phantomsection\label{refs}

\endgroup

\clearpage

\appendix
\makeatletter
\let\apx@seccntformat\@seccntformat
\renewcommand\@seccntformat[1]{\ifstrequal{#1}{section}{\appendixname~}{}\apx@seccntformat{#1}}
\makeatother
\singlespacing
\fontsize{10}{12}\selectfont
\setlength{\parskip}{4pt}
\AtBeginEnvironment{longtable}{\fontsize{10}{12}\selectfont}

\section{Data and Estimation Details}\label{sec-dataappendix}

Table \ref{tbl-crosswalk} supplies the complete item inventory. The
following diagnostics and settings support the data interpretation and
estimation procedures in the main text.

The featured vocabulary retains lemmas appearing in at least 20
documents and no more than 80\% of the processed MathQA corpus. The
stricter thresholds are fixed-design perturbations. The corpus contains
29,837 processed problems and the encoder produces 4,096-dimensional
vectors. The saved item run used fp32, batches of five, and a maximum
length of 32,768 tokens. Item lengths were 25--126 tokens, so no item
was truncated. The encoder revision was not commit-pinned; the original
word-occurrence run's hardware and complete execution settings were not
retained. The archive supplies the resulting contextual-score matrix
rather than the extraction code, so these upstream provenance limits
remain.

\begin{center}
\begingroup\singlespacing
\setlength{\tabcolsep}{2.4pt}
\renewcommand{\arraystretch}{0.95}
\fontsize{9.0}{10.5}\selectfont
\begin{threeparttable}
\captionof{table}{Item crosswalk and response summaries.}
\label{tbl-crosswalk}
\begin{tabularx}{\textwidth}{@{}r@{\hspace{3pt}}l@{\hspace{4pt}}>{\raggedright\arraybackslash}X c c c r r r@{}}
\toprule
No. & ID & Short item label & F & Cont. & Cog. & $N$ & $a$ & $b$ \\
\midrule
1 & M032064 & Divide 560 zeds & CR & Num. & Apply & 1433 & 1.67 & 0.90 \\
2 & M032094 & $4/100+3/1000$ & MC & Num. & Know & 1476 & 1.11 & -0.75 \\
3 & M032166 & Estimate $(7.21\times3.86)/10.09$ & MC & Num. & Know & 1472 & 1.31 & -1.43 \\
4 & M032626 & Prime factors of 36 & MC & Num. & Know & 1462 & 0.96 & -0.73 \\
5 & M032725 & Convert $3\,5/6$ to decimal & CR & Num. & Know & 1417 & 1.68 & 0.77 \\
6 & M042032 & Fraction equivalent to 0.125 & MC & Num. & Know & 1462 & 1.11 & -1.31 \\
7 & M042041 & Original pipe length & MC & Num. & Apply & 1475 & 1.29 & -1.04 \\
8 & M042186 & Next line in a pattern & CR & Num. & Reason & 1458 & 1.33 & -0.39 \\
9 & M052061 & Pack eggs into boxes & CR & Num. & Apply & 1450 & 1.43 & -0.30 \\
10 & M052214 & True number sentence & MC & Num. & Know & 1443 & 0.55 & 1.05 \\
11 & M052216 & Decimal equal to $3/5$ & MC & Num. & Know & 1451 & 1.49 & -1.51 \\
12 & M052228 & Subtract fractions & MC & Num. & Apply & 1444 & 0.95 & 1.04 \\
13 & M052231 & Add 42.65 and 5.748 & CR & Num. & Know & 1457 & 0.57 & -4.00 \\
14 & M032047 & Three consecutive whole numbers & MC & Num. & Apply & 1446 & 0.54 & 1.03 \\
15 & M032295 & Parade with $m$ boys and $n$ girls & MC & Alg. & Know & 1486 & 1.97 & -1.55 \\
16 & M032477 & Taxi cost for $n$ km & MC & Alg. & Know & 1474 & 1.72 & -0.52 \\
17 & M032538 & Find $y$ when $t=9$ & CR & Alg. & Know & 1433 & 1.68 & -0.30 \\
18 & M032673 & $t$ between 6 and 9 & MC & Alg. & Know & 1458 & 1.41 & -0.48 \\
19 & M032738 & Meaning of $xy+1$ & MC & Alg. & Know & 1486 & 1.46 & -1.32 \\
20 & M042077 & Equivalent to $4(3+x)$ & MC & Alg. & Know & 1460 & 1.17 & -0.19 \\
21 & M042086 & Value of $2a+2b+4$ & CR & Alg. & Apply & 1355 & 1.30 & 0.60 \\
22 & M042103 & Solve the inequality & CR & Alg. & Know & 1372 & 2.00 & 0.96 \\
23 & M042198A & Next pattern term & CR & Alg. & Know & 1439 & 0.73 & -3.30 \\
24 & M042198B & Pattern term 100 & CR & Alg. & Reason & 1383 & 0.80 & -0.17 \\
25 & M042198C & Pattern term $n$ & CR & Alg. & Reason & 1295 & 2.33 & 1.36 \\
26 & M042226 & Find $P$ & CR & Alg. & Know & 1430 & 1.43 & -0.94 \\
27 & M042235 & Find $x$ and $y$ & MC & Alg. & Know & 1458 & 1.78 & -0.78 \\
28 & M042236 & Simplify the expression & MC & Alg. & Know & 1451 & 1.12 & -0.41 \\
29 & M042245 & Equation for number pairs & MC & Alg. & Apply & 1420 & 1.24 & 0.41 \\
30 & M052302 & Value of $y$ & MC & Alg. & Know & 1433 & 1.70 & -1.80 \\
31 & M032116 & Square with area 144 cm$^2$ & MC & Geom. & Apply & 1472 & 1.09 & 0.14 \\
32 & M032324 & Distance between midpoints & MC & Geom. & Reason & 1452 & 0.89 & 1.11 \\
33 & M032331 & Clock-hand degrees & MC & Geom. & Apply & 1456 & 0.65 & 2.11 \\
34 & M052084 & Area of a square & MC & Geom. & Apply & 1432 & 1.32 & -0.16 \\
35 & M032132 & Chance of pink candy & MC & Data & Know & 1467 & 0.77 & -0.91 \\
36 & M042169A & Staff mean & CR & Data & Know & 1412 & 1.55 & -0.65 \\
37 & M042169B & Staff median & CR & Data & Know & 1411 & 1.30 & -1.76 \\
38 & M042169C & Change in mean and median & CR & Data & Apply & 1342 & 1.97 & 0.14 \\
39 & M042177 & Regular-size bottles & MC & Data & Apply & 1443 & 1.03 & -0.67 \\
40 & M052429 & Probability of a red marble & MC & Data & Reason & 1432 & 1.49 & -0.71 \\
\bottomrule
\end{tabularx}
\begin{tablenotes}[flushleft]\fontsize{9.0}{10.5}\selectfont
\item Note. F = format; Cont./Cog. = official classifications; $N$ = response count; $a$/$b$ = 2PL discrimination/difficulty; A--C = shared-stem parts. Numbers index later tables.
\end{tablenotes}
\end{threeparttable}
\endgroup
\end{center}

\begin{table}

\caption{\label{tbl-score-diagnostics}Marginal shape and row-profile concentration.}

\centering{

\centering
\begingroup\singlespacing
\setlength{\tabcolsep}{2.8pt}
\renewcommand{\arraystretch}{1.02}
\fontsize{9.0}{10.5}\selectfont
\begin{threeparttable}

\begin{tabular}{@{}lrr@{}}
\multicolumn{3}{@{}l}{\textit{Panel A. Item-column marginals}}\\[-2pt]
\toprule
Data object & Skewness, median (range) & Excess kurtosis, median (range) \\
\midrule
Cosine contextual scores & $.14\;(-.14,\ 1.03)$ & $.69\;(.34,\ 4.29)$ \\
Dot-product contextual scores & $.37\;(-.11,\ 1.22)$ & $.83\;(.24,\ 4.11)$ \\
Binary item responses & $-.45\;(-2.74,\ 1.97)$ & $-1.39\;(-2.00,\ 5.53)$ \\
\bottomrule
\end{tabular}

\vspace{5pt}
\begin{tabular}{@{}lrrrrrr@{}}
\multicolumn{7}{@{}l}{\textit{Panel B. Full word-profile summaries}}\\[-2pt]
\toprule
Profile & Rows & Items & Mean $r$ & $|r|\geq.5$ & EV1 share & PR rank \\
\midrule
Full cosine, raw & 1,864 & 40 & .563 & .675 & .578 & 2.86 \\
Full cosine, standardized & 1,864 & 40 & .001 & .065 & .172 & 13.20 \\
\bottomrule
\end{tabular}

\vspace{5pt}
\begin{tabularx}{\textwidth}{@{}>{\raggedright\arraybackslash}Xrr@{}}
\multicolumn{3}{@{}l}{\textit{Panel C. Matched within-booklet comparison}}\\[-2pt]
\toprule
Summary & Cosine word profiles & Response profiles \\
\midrule
Rows per booklet & 1,864 & 548--739 \\
Items per booklet & 3--16 & 3--16 \\
Mean $r$ & $.001\;(-.000,.003)$ & $-.001\;(-.002,.142)$ \\
$|r|\geq.5$ & $.261\;(.143,.682)$ & $.181\;(.060,.778)$ \\
EV1 share & $.275\;(.201,.584)$ & $.210\;(.114,.736)$ \\
PR rank & $6.00\;(1.95,8.85)$ & $7.53\;(1.64,13.77)$ \\
\bottomrule
\end{tabularx}
\begin{tablenotes}[flushleft]\fontsize{9.0}{10.5}\selectfont
\item Note. Panel A is computed separately for each of 40 columns; response-column sample sizes are 1,295--1,486. In Panels B--C, $r$ correlates two centered row profiles, EV1 is first-eigenvalue share, and PR is participation-ratio rank. Because 75.8\% of focal response cells are structurally missing, Panel C summarizes the eight focal-item booklets: response rows are complete within booklet, cosine profiles use the same item subsets, and entries are medians (ranges). These are measures of profile redundancy, not tests of stochastic row independence or effective sample size.
\end{tablenotes}
\end{threeparttable}
\endgroup

}

\end{table}%

\begin{table}[H]
\centering
\begingroup\footnotesize\singlespacing
\setlength{\tabcolsep}{3.5pt}
\renewcommand{\arraystretch}{1.04}
\caption{Estimation and decision settings.}
\label{tbl-estimation-settings}
\begin{tabularx}{\textwidth}{@{}p{.19\textwidth}X@{}}
\toprule
Stage & Setting \\
\midrule
Step 1 & \texttt{vbpm} 0.9.0; R 4.5.1; seed 2026; maximum 10,000 iterations; convergence tolerance $10^{-4}$ \\
Step-1 prior path & $v_0=.01,.005,.002,.001$; raw PIP threshold .50 \\
Candidate design & 24 specifications and 276 fits; content $K=4$--14 and cognitive $K=3$--14 \\
\shortstack[l]{Count and\\persistence} & 20\% BIC/ELBO paths; direct minimum congruence .85/one, .80/two, and .70/three \\
Step 2 & \texttt{vbpm} 0.9.1; four fixed definitions; oblique and bifactor within each; BIC tie band 2 \\
Text references & \texttt{vbpm} 0.9.1; same prior path; 10,000 iterations per stage, with one 20,000-iteration escalation if needed; tolerance $10^{-4}$ \\
Q benchmark & \texttt{GDINA} 2.9.12 and \texttt{cdmTools} 1.0.6; common sample, HO/saturated distributions, and common Hull-PVAF controls in Section \ref{sec-q-benchmark} \\
Response 2PL & \texttt{mirt} 1.46.1; 61 quadrature points; maximum 1,000 cycles \\
Ancillary HO-CDM & \texttt{GDINA} 2.9.12; exact common $N=5{,}930$; package-enforced three starts; maximum 2,000 iterations; d2-overlap escalation used ten external seeds and maximum 5,000 iterations \\
\bottomrule
\end{tabularx}
\vspace{2pt}\parbox{.94\textwidth}{\footnotesize Note. Each stage uses the stated settings; convergence results are reported with the corresponding findings. Between versions 0.9.0 and 0.9.1, the executable fitting, sweep, matching, persistence, and fit-statistic sources used here are unchanged; the version split does not indicate a different Step-1 estimator.}
\endgroup
\end{table}

Response calibrations use GDINA 2.9.12 and cdmTools 1.0.6, requested
three starts, tolerance .0001, and an initial 2,000-iteration cap. A
nonconverged calibration receives one same-seed rerun with a
5,000-iteration cap. Higher-order attribute slopes are bounded .1--5 and
intercepts -4--4, using 25 normal quadrature points and no structural
prior. Hull uses test-attribute iteration, PVAF, and a 100-iteration
cap; the adapted internal calibration calls preserve its selection
logic. All response fits retain the common 5,930-by-40 matrix without
sampling weights. The saturated \(Q_{\mathrm{i}2}\) and
\(Q_{\mathrm{r}2}\) fits required the longer cap. The complete attempt
and termination records are deposited.

Recorded timings provide partial evidence about computational cost. The
depth-two content/MathQA anchor-only sweep over \(K=4\)--14 required
14.18--287.94 seconds per candidate and 950.05 seconds overall. It
returned BIC and ELBO count readings of nine and eight, respectively
(Table \ref{tbl-step1-census}); the featured depth-four specification
instead returned seven and nine. These timings concern the depth-two
specification, rather than the complete featured analysis. A separate
unanchored PEFA grid on 361 items and 1,825 word rows fitted one
candidate at each even factor count from \(K=2\) to 20, giving ten fits
with times of 36.76--3,192.38 seconds. Its largest recorded vector-heap
and node-heap peaks were 112.9 MB and 88.4 MB, respectively. These are
separate R-memory components, not total R or system-memory use. The
larger-bank grid did not evaluate persistence or select a final
structure. Neither record times the complete anchored two-step procedure
and response analyses; the detailed timings and execution settings
accompany the analysis archive.

\section{Structural Specifications and Checks}\label{sec-step1appendix}

The main text reports the featured direct persistence and depth
comparisons. Table \ref{tbl-step1-census} retains every examined
specification. Full absolute-fit and relative-fit trajectories are
supplied as machine-readable records in the analysis archive, including
the candidate fits and adjacent transitions for every sweep cell. The
featured count-path display below supplies the numerical basis for the
main text's count-versus-persistence example.

\begin{table}

\caption{\label{tbl-step1-census}Complete Step-1 persistence results.}

\centering{

\centering
\begingroup\singlespacing
\setlength{\tabcolsep}{2.5pt}
\renewcommand{\arraystretch}{0.96}
\fontsize{9.0}{10.5}\selectfont
\begin{threeparttable}

\begin{tabular}{@{}r c c c r r c c c@{}}
\toprule
Depth & Backbone & Corpus & Spec. & $K_{\mathrm{BIC},20\%}$ & $K_{\mathrm{ELBO},20\%}$ & .85/one & .80/two & .70/three \\
\midrule
2&Cogn.&E&AZ&7&9&--/L3&--/L3&--/L3\\
2&Cogn.&E&AO&7&9&3/L2&3/L2&3/L2\\
2&Cogn.&M&AZ&8&8&6/L2&--/L3&--/L3\\
2&Cogn.&M&AO&7&9&5/L2&3/L2&3/L2\\
2&Cont.&E&AZ&10&10&7/L2&--/L3&--/L3\\
2&Cont.&E&AO&8&9&5/L2&--/L3&--/L3\\
2&Cont.&M&AZ&7&8&--/L3&--/L3&--/L3\\
2&Cont.&M&AO&9&8&7/L2&7/L2&5/L2\\
4&Cogn.&E&AZ&6&7&--/L3&--/L3&--/L3\\
4&Cogn.&E&AO&7&9&4/L2&--/L3&4/L2\\
4&Cogn.&M&AZ&5&8&5/L1&--/L3&--/L3\\
4&Cogn.&M&AO&7&8&--/L3&4/L2&--/L3\\
4&Cont.&E&AZ&8&9&4/L2&--/L3&--/L3\\
4&Cont.&E&AO&8&9&4/L2&4/L2&4/L2\\
4&Cont.&M&AZ&8&9&--/L3&--/L3&--/L3\\
4&Cont.&M&AO&7&9&7/L1&7/L1&7/L1\\
6&Cogn.&E&AZ&7&7&--/L3&--/L3&--/L3\\
6&Cogn.&E&AO&5&9&3/L2&4/L2&3/L2\\
6&Cogn.&M&AZ&8&8&--/L3&--/L3&--/L3\\
6&Cogn.&M&AO&8&8&5/L2&4/L2&--/L3\\
6&Cont.&E&AZ&8&9&--/L3&--/L3&--/L3\\
6&Cont.&E&AO&8&9&7/L2&6/L2&5/L2\\
6&Cont.&M&AZ&7&9&--/L3&--/L3&--/L3\\
6&Cont.&M&AO&5&9&5/L1&--/L3&--/L3\\
\bottomrule
\end{tabular}
\begin{tablenotes}[flushleft]\fontsize{9.0}{10.5}\selectfont
\item Note. Each profile cell gives $K_p$/layer; a dash denotes valid absence. The $K_{\cdot,20\%}$ columns report counts returned by the separately normalized 20\%-gain paths, not raw optima. Cogn. = cognitive backbone; Cont. = content backbone; M = MathQA; E = nested MathQA-plus-GSM8K extension; AO = anchor only; AZ = anchor zero. The featured row is depth 4, content, MathQA, AO.
\end{tablenotes}
\end{threeparttable}
\endgroup

}

\end{table}%

\begin{table}

\caption{\label{tbl-trajectory}Featured Step-1 count path and adjacent comparisons.}

\centering{

\centering
\begingroup\footnotesize\singlespacing
\setlength{\tabcolsep}{4.0pt}
\renewcommand{\arraystretch}{0.98}

\begin{tabular}{@{}lrrrrc@{}}
\toprule
Step & ELBO gain (\%) & BIC gain (\%) & Adjacent $\phi_{\min}$ & RMSD$_{\max}$ & Collision \\
\midrule
$4\rightarrow5$   & 100.0 & 100.0 & .986 & .096 & No \\
$5\rightarrow6$   & 56.3  & 56.9  & .451 & .227 & No \\
$6\rightarrow7$   & 48.9  & 36.4  & .928 & .160 & No \\
$7\rightarrow8$   & 29.7  & 1.6   & .900 & .181 & No \\
$8\rightarrow9$   & 20.7  & 30.6  & .742 & .228 & No \\
$9\rightarrow10$  & 14.7  & -17.9 & .589 & .213 & No \\
$10\rightarrow11$ & 14.0  & 12.3  & .355 & .353 & Yes \\
$11\rightarrow12$ & 14.1  & -29.1 & .293 & .277 & No \\
$12\rightarrow13$ & 7.6   & -45.3 & .405 & .313 & Yes \\
$13\rightarrow14$ & 7.3   & -19.6 & .019 & .340 & Yes \\
\bottomrule
\end{tabular}
\vspace{2pt}\parbox{.95\textwidth}{\footnotesize Note. MathQA corpus, content backbone, anchor-only specification, depth four. Gains are normalized by the largest positive gain on each path; $\mathcal{K}_{20\%}=\{7,9\}$. The adjacent congruence and root-mean-square-difference (RMSD) columns diagnose each displayed move but do not establish persistence.}
\endgroup

}

\end{table}%

\subsection{Framework anchors}\label{framework-anchors}

Table \ref{tbl-r51-anchor-ladder} lists imposed anchors by nominal
depth. Item indices refer to the main item crosswalk. The same
assignments apply across both corpora and both anchor modes; geometry
has only four items and reasoning five, so their depth-six sets are
capped. These entries are imposed associations, not estimated high-PIP
discoveries.

\begin{longtable}[]{@{}
  >{\raggedright\arraybackslash}p{(\linewidth - 6\tabcolsep) * \real{0.2500}}
  >{\raggedright\arraybackslash}p{(\linewidth - 6\tabcolsep) * \real{0.2500}}
  >{\raggedright\arraybackslash}p{(\linewidth - 6\tabcolsep) * \real{0.2500}}
  >{\raggedright\arraybackslash}p{(\linewidth - 6\tabcolsep) * \real{0.2500}}@{}}
\caption{Complete inherited anchor
assignments.}\label{tbl-r51-anchor-ladder}\tabularnewline
\toprule\noalign{}
\begin{minipage}[b]{\linewidth}\raggedright
Framework and domain
\end{minipage} & \begin{minipage}[b]{\linewidth}\raggedright
Depth 2
\end{minipage} & \begin{minipage}[b]{\linewidth}\raggedright
Depth 4
\end{minipage} & \begin{minipage}[b]{\linewidth}\raggedright
Depth 6
\end{minipage} \\
\midrule\noalign{}
\endfirsthead
\toprule\noalign{}
\begin{minipage}[b]{\linewidth}\raggedright
Framework and domain
\end{minipage} & \begin{minipage}[b]{\linewidth}\raggedright
Depth 2
\end{minipage} & \begin{minipage}[b]{\linewidth}\raggedright
Depth 4
\end{minipage} & \begin{minipage}[b]{\linewidth}\raggedright
Depth 6
\end{minipage} \\
\midrule\noalign{}
\endhead
\bottomrule\noalign{}
\endlastfoot
Content: algebra & 22, 28 & 20, 22, 27, 28 & 19, 20, 22, 25, 27, 28 \\
Content: data and chance & 35, 40 & 35, 36, 39, 40 & 35, 36, 37, 38, 39,
40 \\
Content: geometry & 31, 32 & 31, 32, 33, 34 & 31, 32, 33, 34 (cap) \\
Content: number & 2, 6 & 2, 4, 5, 6 & 2, 3, 4, 5, 6, 12 \\
Cognitive: applying & 7, 31 & 7, 21, 31, 39 & 7, 9, 21, 31, 33, 39 \\
Cognitive: knowing & 13, 19 & 4, 13, 19, 36 & 4, 13, 19, 30, 35, 36 \\
Cognitive: reasoning & 8, 40 & 8, 25, 32, 40 & 8, 24, 25, 32, 40
(cap) \\
\end{longtable}

\subsection{Added-group markers}\label{added-group-markers}

Table~\ref{tbl-marker-assignments} makes the two constructors
reproducible. The overlap rule ranks items within each added Step-1
column by absolute loading, breaks exact ties by item number, and
permits reuse. The globally disjoint rule solves one
maximum-total-absolute-loading assignment over all added-marker slots;
inherited anchors are ineligible and each item can occupy only one slot.
A deterministic item-index perturbation resolves exact ties in favor of
lower-numbered eligible items.

\begin{longtable}[]{@{}
  >{\raggedright\arraybackslash}p{(\linewidth - 8\tabcolsep) * \real{0.1875}}
  >{\raggedleft\arraybackslash}p{(\linewidth - 8\tabcolsep) * \real{0.2500}}
  >{\raggedright\arraybackslash}p{(\linewidth - 8\tabcolsep) * \real{0.1875}}
  >{\raggedright\arraybackslash}p{(\linewidth - 8\tabcolsep) * \real{0.1875}}
  >{\raggedright\arraybackslash}p{(\linewidth - 8\tabcolsep) * \real{0.1875}}@{}}
\caption{Added-column
markers.}\label{tbl-marker-assignments}\tabularnewline
\toprule\noalign{}
\begin{minipage}[b]{\linewidth}\raggedright
Definition
\end{minipage} & \begin{minipage}[b]{\linewidth}\raggedleft
Backbone markers per S1--S4
\end{minipage} & \begin{minipage}[b]{\linewidth}\raggedright
S5 markers
\end{minipage} & \begin{minipage}[b]{\linewidth}\raggedright
S6 markers
\end{minipage} & \begin{minipage}[b]{\linewidth}\raggedright
S7 markers
\end{minipage} \\
\midrule\noalign{}
\endfirsthead
\toprule\noalign{}
\begin{minipage}[b]{\linewidth}\raggedright
Definition
\end{minipage} & \begin{minipage}[b]{\linewidth}\raggedleft
Backbone markers per S1--S4
\end{minipage} & \begin{minipage}[b]{\linewidth}\raggedright
S5 markers
\end{minipage} & \begin{minipage}[b]{\linewidth}\raggedright
S6 markers
\end{minipage} & \begin{minipage}[b]{\linewidth}\raggedright
S7 markers
\end{minipage} \\
\midrule\noalign{}
\endhead
\bottomrule\noalign{}
\endlastfoot
Depth-4 overlap & 4 & 9, 15, 35, 40 & 15, 36, 37, 38 & 7, 9, 13, 33 \\
Depth-4 disjoint & 4 & 9, 16, 17, 23 & 1, 15, 37, 38 & 7, 8, 13, 18 \\
Depth-2 overlap & 2 & 36, 38 & 13, 33 & 35, 40 \\
Depth-2 disjoint & 2 & 36, 38 & 13, 33 & 9, 15 \\
\end{longtable}

\begin{minipage}{\linewidth}
\footnotesize\textit{Note.} Item numbers refer to Table \ref{tbl-crosswalk}. Global disjointness applies to marker assignment, not fitted loading support.
\end{minipage}

\subsection{Alignment and identification
checks}\label{alignment-and-identification-checks}

For the adopted hard-selected covariance, the parameter count is
\(173=40+93+40\): general loadings, selected group loadings, and
uniquenesses. The covariance Jacobian has 820 rows and 173 columns and
full numerical rank at tolerance \(10^{-6}\) times the largest singular
value. All definition-specific Step-2 fits have equal nominal and rank
counts. These are local numerical checks of the selected covariance, not
global identification proofs.

Direct observed-minus-implied covariance residuals provide another view
of the text approximation (Table \ref{tbl-text-residuals}). The
continuous unthresholded plug-in covariance has smaller residuals than
the hard-selected covariance. The package standardized root mean square
residual (SRMR) includes diagonal entries, whereas the summaries below
use unique off-diagonal pairs. These summaries expose approximation
error; they do not establish an independent-row sampling model or
justify conventional respondent-sample cutoffs.

\begin{longtable}[]{@{}
  >{\raggedright\arraybackslash}p{(\linewidth - 4\tabcolsep) * \real{0.2727}}
  >{\raggedleft\arraybackslash}p{(\linewidth - 4\tabcolsep) * \real{0.3636}}
  >{\raggedleft\arraybackslash}p{(\linewidth - 4\tabcolsep) * \real{0.3636}}@{}}
\caption{Text-covariance approximation
residuals.}\label{tbl-text-residuals}\tabularnewline
\toprule\noalign{}
\begin{minipage}[b]{\linewidth}\raggedright
Off-diagonal residual summary
\end{minipage} & \begin{minipage}[b]{\linewidth}\raggedleft
Hard-selected covariance
\end{minipage} & \begin{minipage}[b]{\linewidth}\raggedleft
Continuous covariance
\end{minipage} \\
\midrule\noalign{}
\endfirsthead
\toprule\noalign{}
\begin{minipage}[b]{\linewidth}\raggedright
Off-diagonal residual summary
\end{minipage} & \begin{minipage}[b]{\linewidth}\raggedleft
Hard-selected covariance
\end{minipage} & \begin{minipage}[b]{\linewidth}\raggedleft
Continuous covariance
\end{minipage} \\
\midrule\noalign{}
\endhead
\bottomrule\noalign{}
\endlastfoot
Root mean square & .0284 & .0150 \\
Mean absolute & .0219 & .0105 \\
Maximum absolute & .1066 & .0824 \\
\end{longtable}

The summaries use all 780 unique item pairs. The hard-selected
covariance is defined in Equation \ref{eq-hard-covariance}; its package
SRMR is .0278.

\subsection{Descriptive EFA and embedding-clustering
references}\label{descriptive-efa-and-embedding-clustering-references}

Table \ref{tbl-r52-comparators} summarizes alternative representations.
The EFA uses maximum likelihood with oblimin rotation on the same
standardized 1,864-by-40 contextual-score matrix. Its nine-factor count
comes from the working-BIC gain reading of the depth-two content/MathQA
anchor-only sweep, so it is not matched in count to the seven-group
model. The \(k\)-means and Ward clusterings use L2-normalized raw item
embeddings at \(k=4\), the number of official content domains, and
\(k=8\), the nine-factor reference count less its one broad elevation
column. Their exclusive clusters differ from overlapping semantic
memberships; Cramér's \(V\) and the adjusted Rand index (ARI) describe
alignment with the TIMSS official content framework. These summaries
establish neither computational savings nor diagnostic benefit, and
their fit criteria are not compared with the contextual-score working
BIC.

\begin{table}

\caption{\label{tbl-r52-comparators}Descriptive summaries of alternative item representations.}

\centering{

\centering
\begingroup\small\singlespacing
\setlength{\tabcolsep}{6pt}

\begin{tabular}{@{}lrrrr@{}}
\toprule
\multicolumn{5}{@{}l}{\textit{Panel A. Nine-factor EFA projection}}\\[2pt]
Threshold & Memberships & Density & Unassigned & Multiple\\
\midrule
$|\lambda|>.30$ & 44 & .1222 & 0 & 4\\
\midrule
\multicolumn{5}{@{}l}{\textit{Panel B. Raw-embedding cluster alignment with content}}\\[2pt]
Method & Clusters & Cramér's $V$ & ARI & \\
\midrule
$k$-means & 4 & .5566 & .2327 & \\
$k$-means & 8 & .8373 & .3170 & \\
Ward & 4 & .5638 & .2461 & \\
Ward & 8 & .8344 & .3026 & \\
\bottomrule
\end{tabular}
\par\vspace{3pt}\parbox{.94\textwidth}{\footnotesize Note. All summaries use 40 items. Density is the proportion of selected cells in the 40-by-9 EFA map. Unassigned and Multiple count items with zero and more than one selected loading, respectively. Clustering alignment concerns official content labels, not student responses.}
\endgroup

}

\end{table}%

\clearpage

\section{Complete Semantic Maps}\label{sec-mapappendix}

Signed loadings (Table \ref{tbl-full-loadings}) and raw group PIPs
(Table \ref{tbl-full-pips}) jointly define the graded semantic
item--factor map; the thresholded binary projection is supplied as a
machine-readable Q in the analysis archive. A group cell is present when
imposed or when its raw PIP is at least .50; the general column is
excluded from the diagnostic Q. Of the 93 selected group cells, 19 have
negative loadings: 17 are freely selected and two are imposed. Their
inclusion illustrates the loss of direction when semantic associations
are converted to positive binary entries.

\begin{table}

\caption{\label{tbl-full-loadings}Adopted bifactor loadings.}

\centering{

\centering
\begingroup\singlespacing
\setlength{\tabcolsep}{3.0pt}
\renewcommand{\arraystretch}{0.95}
\fontsize{9.0}{10.5}\selectfont

\begin{tabular}{@{}r r r r r r r r r@{}}
\toprule
No. & G & S1 & S2 & S3 & S4 & S5 & S6 & S7 \\
\midrule
1 & $\mathbf{.821}$ & $-.052$ & $.063$ & $-.152$ & $-.014$ & $-.196$ & $.159$ & $.004$ \\
2 & $\mathbf{.918}$ & $.060$ & $-.041$ & $-.076$ & $.146$ & $-.105$ & $-.009$ & $.044$ \\
3 & $\mathbf{.902}$ & $.121$ & $-.060$ & $.080$ & $.159$ & $-.009$ & $.027$ & $.056$ \\
4 & $\mathbf{.846}$ & $.229$ & $.102$ & $.146$ & $.124$ & $.112$ & $.044$ & $-.037$ \\
5 & $\mathbf{.892}$ & $-.021$ & $-.045$ & $.066$ & $.188$ & $-.057$ & $-.037$ & $.194$ \\
6 & $\mathbf{.924}$ & $-.031$ & $.056$ & $.020$ & $.271$ & $-.029$ & $-.037$ & $-.002$ \\
7 & $\mathbf{.788}$ & $-.033$ & $.023$ & $.209$ & $.042$ & $-.182$ & $-.016$ & $.212$ \\
8 & $\mathbf{.789}$ & $.298$ & $.022$ & $-.052$ & $-.050$ & $\mathbf{.331}$ & $-.028$ & $.203$ \\
9 & $\mathbf{.779}$ & $.136$ & $.287$ & $-.016$ & $-.003$ & $-.106$ & $.176$ & $.240$ \\
10 & $\mathbf{.919}$ & $.063$ & $.003$ & $-.030$ & $.173$ & $.011$ & $-.008$ & $-.194$ \\
11 & $\mathbf{.916}$ & $.000$ & $.056$ & $.065$ & $\mathbf{.338}$ & $.006$ & $.039$ & $.027$ \\
12 & $\mathbf{.904}$ & $.124$ & $.033$ & $.055$ & $.139$ & $.137$ & $-.009$ & $-.020$ \\
13 & $\mathbf{.854}$ & $.120$ & $-.109$ & $-.004$ & $-.048$ & $-.167$ & $.028$ & $.283$ \\
14 & $\mathbf{.795}$ & $.233$ & $-.012$ & $.008$ & $.021$ & $\mathbf{.332}$ & $.100$ & $.071$ \\
15 & $\mathbf{.731}$ & $.180$ & $\mathbf{.327}$ & $-.028$ & $-.016$ & $-.002$ & $\mathbf{.306}$ & $-.001$ \\
16 & $\mathbf{.691}$ & $.023$ & $-.146$ & $-.009$ & $-.187$ & $-.182$ & $.072$ & $.010$ \\
17 & $\mathbf{.893}$ & $.070$ & $-.160$ & $.007$ & $-.053$ & $.001$ & $-.043$ & $-.005$ \\
18 & $\mathbf{.810}$ & $.281$ & $.010$ & $-.017$ & $.005$ & $.194$ & $.135$ & $.245$ \\
19 & $\mathbf{.796}$ & $\mathbf{.442}$ & $-.087$ & $-.001$ & $-.129$ & $.182$ & $-.024$ & $-.015$ \\
20 & $\mathbf{.877}$ & $\mathbf{.368}$ & $-.070$ & $.072$ & $.007$ & $-.021$ & $.006$ & $-.030$ \\
21 & $\mathbf{.842}$ & $\mathbf{.397}$ & $-.069$ & $.004$ & $-.144$ & $-.003$ & $-.033$ & $.053$ \\
22 & $\mathbf{.789}$ & $\mathbf{.439}$ & $.009$ & $.008$ & $.001$ & $-.003$ & $-.053$ & $.226$ \\
23 & $\mathbf{.929}$ & $.014$ & $.106$ & $-.037$ & $.018$ & $.259$ & $.030$ & $.152$ \\
24 & $\mathbf{.914}$ & $-.010$ & $.035$ & $-.048$ & $-.035$ & $\mathbf{.361}$ & $.033$ & $-.006$ \\
25 & $\mathbf{.864}$ & $.022$ & $.055$ & $.000$ & $.002$ & $\mathbf{.472}$ & $.059$ & $-.034$ \\
26 & $\mathbf{.888}$ & $.292$ & $-.008$ & $-.050$ & $-.048$ & $.048$ & $.031$ & $-.001$ \\
27 & $\mathbf{.738}$ & $\mathbf{.569}$ & $.001$ & $-.024$ & $.004$ & $-.043$ & $-.002$ & $-.071$ \\
28 & $\mathbf{.842}$ & $\mathbf{.404}$ & $.042$ & $.029$ & $-.045$ & $.019$ & $-.047$ & $-.020$ \\
29 & $\mathbf{.720}$ & $\mathbf{.607}$ & $.007$ & $-.013$ & $.030$ & $.153$ & $-.045$ & $.009$ \\
30 & $\mathbf{.907}$ & $.271$ & $-.018$ & $.041$ & $.047$ & $.037$ & $.110$ & $-.013$ \\
31 & $\mathbf{.671}$ & $.013$ & $-.061$ & $\mathbf{.717}$ & $.004$ & $-.028$ & $-.031$ & $-.021$ \\
32 & $\mathbf{.621}$ & $\mathbf{.351}$ & $-.033$ & $.270$ & $.052$ & $.086$ & $-.006$ & $.274$ \\
33 & $\mathbf{.617}$ & $-.026$ & $-.026$ & $.147$ & $.027$ & $-.033$ & $-.002$ & $\mathbf{.404}$ \\
34 & $\mathbf{.730}$ & $-.018$ & $.029$ & $\mathbf{.661}$ & $.009$ & $-.033$ & $-.001$ & $.026$ \\
35 & $\mathbf{.626}$ & $-.004$ & $\mathbf{.720}$ & $-.161$ & $-.020$ & $-.011$ & $.010$ & $.035$ \\
36 & $\mathbf{.788}$ & $-.013$ & $.142$ & $-.049$ & $.001$ & $-.031$ & $\mathbf{.580}$ & $.003$ \\
37 & $\mathbf{.757}$ & $-.020$ & $.174$ & $.028$ & $.045$ & $.069$ & $\mathbf{.603}$ & $-.011$ \\
38 & $\mathbf{.658}$ & $-.063$ & $.146$ & $-.124$ & $-.061$ & $.009$ & $\mathbf{.615}$ & $-.022$ \\
39 & $\mathbf{.769}$ & $-.005$ & $.171$ & $-.199$ & $.036$ & $\mathbf{-.319}$ & $.177$ & $-.150$ \\
40 & $\mathbf{.705}$ & $-.040$ & $\mathbf{.640}$ & $-.022$ & $.054$ & $.059$ & $-.005$ & $-.055$ \\
\bottomrule
\end{tabular}

\par\vspace{1pt}\parbox{0.95\textwidth}{\fontsize{9.0}{10.5}\selectfont Note. Bold marks $|\lambda|>.30$ as a visual aid only. Numbers refer to Table \ref{tbl-crosswalk}. Columns use the Methods reporting orientation; signs are relative. G is not a diagnostic target.}
\endgroup

}

\end{table}%

\begin{table}

\caption{\label{tbl-full-pips}Graded semantic item--factor map.}

\centering{

\centering
\begingroup\singlespacing
\setlength{\tabcolsep}{3.2pt}
\renewcommand{\arraystretch}{0.95}
\fontsize{9.0}{10.5}\selectfont

\begin{tabular}{@{}r r r r r r r r r@{}}
\toprule
No. & G & S1 & S2 & S3 & S4 & S5 & S6 & S7 \\
\midrule
1 & $1.000^{*}$ & $.047$ & $.088$ & $\mathbf{.999}$ & $.014$ & $\mathbf{1.000}$ & $1.000^{*}$ & $.013$ \\
2 & $1.000^{*}$ & $.071$ & $.029$ & $.185$ & $1.000^{*}$ & $\mathbf{.756}$ & $.013$ & $.032$ \\
3 & $1.000^{*}$ & $\mathbf{.950}$ & $.070$ & $.230$ & $\mathbf{1.000}$ & $.013$ & $.018$ & $.058$ \\
4 & $1.000^{*}$ & $\mathbf{1.000}$ & $\mathbf{.695}$ & $\mathbf{.998}$ & $1.000^{*}$ & $\mathbf{.867}$ & $.032$ & $.025$ \\
5 & $1.000^{*}$ & $.015$ & $.033$ & $.102$ & $1.000^{*}$ & $.060$ & $.024$ & $\mathbf{1.000}$ \\
6 & $1.000^{*}$ & $.020$ & $.056$ & $.015$ & $1.000^{*}$ & $.019$ & $.024$ & $.012$ \\
7 & $1.000^{*}$ & $.022$ & $.017$ & $\mathbf{1.000}$ & $.031$ & $\mathbf{1.000}$ & $.015$ & $1.000^{*}$ \\
8 & $1.000^{*}$ & $\mathbf{1.000}$ & $.016$ & $.045$ & $.042$ & $\mathbf{1.000}$ & $.018$ & $1.000^{*}$ \\
9 & $1.000^{*}$ & $\mathbf{.993}$ & $\mathbf{1.000}$ & $.014$ & $.013$ & $1.000^{*}$ & $\mathbf{1.000}$ & $\mathbf{1.000}$ \\
10 & $1.000^{*}$ & $.081$ & $.012$ & $.019$ & $\mathbf{1.000}$ & $.013$ & $.013$ & $\mathbf{1.000}$ \\
11 & $1.000^{*}$ & $.012$ & $.056$ & $.094$ & $\mathbf{1.000}$ & $.012$ & $.025$ & $.017$ \\
12 & $1.000^{*}$ & $\mathbf{.965}$ & $.021$ & $.055$ & $\mathbf{.995}$ & $\mathbf{.993}$ & $.013$ & $.015$ \\
13 & $1.000^{*}$ & $\mathbf{.946}$ & $\mathbf{.823}$ & $.013$ & $.039$ & $\mathbf{1.000}$ & $.018$ & $1.000^{*}$ \\
14 & $1.000^{*}$ & $\mathbf{1.000}$ & $.013$ & $.013$ & $.016$ & $\mathbf{1.000}$ & $\mathbf{.662}$ & $.133$ \\
15 & $1.000^{*}$ & $\mathbf{1.000}$ & $\mathbf{1.000}$ & $.019$ & $.015$ & $.013$ & $1.000^{*}$ & $.013$ \\
16 & $1.000^{*}$ & $.017$ & $\mathbf{.998}$ & $.014$ & $\mathbf{1.000}$ & $1.000^{*}$ & $.151$ & $.014$ \\
17 & $1.000^{*}$ & $.125$ & $\mathbf{1.000}$ & $.013$ & $.050$ & $1.000^{*}$ & $.030$ & $.013$ \\
18 & $1.000^{*}$ & $\mathbf{1.000}$ & $.013$ & $.014$ & $.013$ & $\mathbf{1.000}$ & $\mathbf{.991}$ & $1.000^{*}$ \\
19 & $1.000^{*}$ & $\mathbf{1.000}$ & $.361$ & $.012$ & $\mathbf{.980}$ & $\mathbf{1.000}$ & $.016$ & $.014$ \\
20 & $1.000^{*}$ & $1.000^{*}$ & $.129$ & $.142$ & $.013$ & $.015$ & $.013$ & $.019$ \\
21 & $1.000^{*}$ & $\mathbf{1.000}$ & $.116$ & $.012$ & $\mathbf{.998}$ & $.012$ & $.021$ & $.048$ \\
22 & $1.000^{*}$ & $1.000^{*}$ & $.013$ & $.013$ & $.012$ & $.012$ & $.050$ & $\mathbf{1.000}$ \\
23 & $1.000^{*}$ & $.013$ & $\mathbf{.762}$ & $.024$ & $.014$ & $1.000^{*}$ & $.019$ & $\mathbf{.999}$ \\
24 & $1.000^{*}$ & $.013$ & $.022$ & $.037$ & $.022$ & $\mathbf{1.000}$ & $.020$ & $.012$ \\
25 & $1.000^{*}$ & $.015$ & $.053$ & $.012$ & $.012$ & $\mathbf{1.000}$ & $.065$ & $.022$ \\
26 & $1.000^{*}$ & $\mathbf{1.000}$ & $.013$ & $.041$ & $.039$ & $.039$ & $.020$ & $.012$ \\
27 & $1.000^{*}$ & $1.000^{*}$ & $.012$ & $.016$ & $.012$ & $.030$ & $.012$ & $.136$ \\
28 & $1.000^{*}$ & $1.000^{*}$ & $.030$ & $.019$ & $.033$ & $.015$ & $.036$ & $.015$ \\
29 & $1.000^{*}$ & $\mathbf{1.000}$ & $.013$ & $.013$ & $.019$ & $\mathbf{.999}$ & $.033$ & $.013$ \\
30 & $1.000^{*}$ & $\mathbf{1.000}$ & $.014$ & $.028$ & $.037$ & $.024$ & $\mathbf{.840}$ & $.013$ \\
31 & $1.000^{*}$ & $.013$ & $.075$ & $1.000^{*}$ & $.012$ & $.018$ & $.019$ & $.015$ \\
32 & $1.000^{*}$ & $\mathbf{1.000}$ & $.022$ & $1.000^{*}$ & $.048$ & $.355$ & $.013$ & $\mathbf{1.000}$ \\
33 & $1.000^{*}$ & $.019$ & $.019$ & $1.000^{*}$ & $.019$ & $.023$ & $.013$ & $\mathbf{1.000}$ \\
34 & $1.000^{*}$ & $.014$ & $.018$ & $1.000^{*}$ & $.013$ & $.021$ & $.012$ & $.017$ \\
35 & $1.000^{*}$ & $.012$ & $1.000^{*}$ & $\mathbf{1.000}$ & $.015$ & $.013$ & $.013$ & $.022$ \\
36 & $1.000^{*}$ & $.013$ & $1.000^{*}$ & $.040$ & $.012$ & $.019$ & $\mathbf{1.000}$ & $.012$ \\
37 & $1.000^{*}$ & $.015$ & $\mathbf{1.000}$ & $.018$ & $.033$ & $.116$ & $1.000^{*}$ & $.013$ \\
38 & $1.000^{*}$ & $.085$ & $\mathbf{.998}$ & $\mathbf{.965}$ & $.077$ & $.013$ & $1.000^{*}$ & $.016$ \\
39 & $1.000^{*}$ & $.013$ & $1.000^{*}$ & $\mathbf{1.000}$ & $.024$ & $\mathbf{1.000}$ & $\mathbf{1.000}$ & $\mathbf{.999}$ \\
40 & $1.000^{*}$ & $.027$ & $1.000^{*}$ & $.016$ & $.051$ & $.065$ & $.012$ & $.053$ \\
\bottomrule
\end{tabular}

\par\vspace{2pt}\parbox{0.95\textwidth}{\footnotesize Note. Item numbers refer to Table \ref{tbl-crosswalk}. G is fixed present for every item. For a free group cell, the raw PIP is variational posterior support for the slab component under the adopted model, design, and prior; it is not the probability that a cognitive attribute is required. $^{*}$ marks imposed presence: the all-item G column or a group-marker cell. Bold marks a free cell with PIP $\ge .50$. The 28 fixed and 65 selected free group cells produce the deposited binary projection.}
\endgroup

}

\end{table}%

\clearpage

\section{Response Analyses and Diagnostics}\label{sec-responseappendix}

\subsection{Q constructions and empirical
revisions}\label{q-constructions-and-empirical-revisions}

Table \ref{tbl-r51-initial-pcfa} distinguishes the 40 imposed official
memberships of \(Q_{\mathrm{i}3}\) from the 47 memberships selected from
contextual scores. No student responses enter this initial projection.

\begin{table}

\caption{\label{tbl-r51-initial-pcfa}Initial framework-guided PCFA Q-matrix, with imposed and selected cells distinguished.}

\centering{

\centering
\begingroup\singlespacing
\fontsize{9.0}{10.5}\selectfont
\setlength{\tabcolsep}{4pt}
\renewcommand{\arraystretch}{1.0}

\begin{tabular}{@{}rlrrrr@{\hspace{16pt}}rlrrrr@{}}
\toprule
No. & Item & Alg. & Data & Geom. & Num. & No. & Item & Alg. & Data & Geom. & Num. \\
\midrule
1 & M032064 & 0 & $\mathbf{1}$ & $\mathbf{1}^{\dagger}$ & $1^*$ & 21 & M042086 & $1^*$ & 0 & 0 & $\mathbf{1}^{\dagger}$ \\
2 & M032094 & $\mathbf{1}$ & 0 & $\mathbf{1}^{\dagger}$ & $1^*$ & 22 & M042103 & $1^*$ & 0 & 0 & 0 \\
3 & M032166 & $\mathbf{1}$ & 0 & $\mathbf{1}$ & $1^*$ & 23 & M042198A & $1^*$ & $\mathbf{1}$ & 0 & $\mathbf{1}$ \\
4 & M032626 & $\mathbf{1}$ & $\mathbf{1}$ & $\mathbf{1}$ & $1^*$ & 24 & M042198B & $1^*$ & $\mathbf{1}$ & 0 & $\mathbf{1}$ \\
5 & M032725 & 0 & 0 & 0 & $1^*$ & 25 & M042198C & $1^*$ & $\mathbf{1}$ & 0 & $\mathbf{1}$ \\
6 & M042032 & 0 & 0 & 0 & $1^*$ & 26 & M042226 & $1^*$ & $\mathbf{1}$ & 0 & 0 \\
7 & M042041 & 0 & 0 & $\mathbf{1}$ & $1^*$ & 27 & M042235 & $1^*$ & 0 & 0 & $\mathbf{1}^{\dagger}$ \\
8 & M042186 & $\mathbf{1}$ & 0 & 0 & $1^*$ & 28 & M042236 & $1^*$ & 0 & 0 & 0 \\
9 & M052061 & $\mathbf{1}$ & $\mathbf{1}$ & 0 & $1^*$ & 29 & M042245 & $1^*$ & 0 & 0 & $\mathbf{1}^{\dagger}$ \\
10 & M052214 & $\mathbf{1}$ & 0 & 0 & $1^*$ & 30 & M052302 & $1^*$ & $\mathbf{1}$ & 0 & $\mathbf{1}$ \\
11 & M052216 & 0 & 0 & 0 & $1^*$ & 31 & M032116 & 0 & 0 & $1^*$ & 0 \\
12 & M052228 & $\mathbf{1}$ & 0 & 0 & $1^*$ & 32 & M032324 & $\mathbf{1}$ & 0 & $1^*$ & $\mathbf{1}^{\dagger}$ \\
13 & M052231 & $\mathbf{1}$ & 0 & 0 & $1^*$ & 33 & M032331 & 0 & 0 & $1^*$ & $\mathbf{1}$ \\
14 & M032047 & $\mathbf{1}$ & $\mathbf{1}$ & 0 & $1^*$ & 34 & M052084 & 0 & 0 & $1^*$ & $\mathbf{1}$ \\
15 & M032295 & $1^*$ & $\mathbf{1}$ & 0 & 0 & 35 & M032132 & 0 & $1^*$ & $\mathbf{1}^{\dagger}$ & $\mathbf{1}$ \\
16 & M032477 & $1^*$ & $\mathbf{1}$ & 0 & $\mathbf{1}$ & 36 & M042169A & 0 & $1^*$ & 0 & 0 \\
17 & M032538 & $1^*$ & 0 & 0 & $\mathbf{1}$ & 37 & M042169B & 0 & $1^*$ & $\mathbf{1}$ & 0 \\
18 & M032673 & $1^*$ & $\mathbf{1}$ & 0 & 0 & 38 & M042169C & 0 & $1^*$ & 0 & 0 \\
19 & M032738 & $1^*$ & 0 & 0 & $\mathbf{1}^{\dagger}$ & 39 & M042177 & 0 & $1^*$ & $\mathbf{1}^{\dagger}$ & $\mathbf{1}$ \\
20 & M042077 & $1^*$ & 0 & $\mathbf{1}$ & 0 & 40 & M052429 & 0 & $1^*$ & 0 & $\mathbf{1}$ \\
\bottomrule
\end{tabular}
\par\vspace{2pt}\parbox{.96\textwidth}{\footnotesize Note. Columns are algebra, data and chance, geometry, and number. $1^*$ is an imposed official membership and has no estimated PIP. Bold 1 marks a selected free cell (raw PIP $\ge .50$); a dagger flags a negative signed loading (nine cells). Zero denotes nonselection under this fit, not absence of a cognitive requirement. The 87 memberships comprise 40 imposed and 47 selected additions. Signed loadings and PIPs are deposited.}
\endgroup

}

\end{table}%

Table \ref{tbl-r51-hull-changes} lists every Hull edit relative to its
own initial Q. The PCFA initial additions are separately deposited with
signed loadings and PIPs; all 40 official memberships were retained
before response revision.

\begin{table}

\caption{\label{tbl-r51-hull-changes}All Hull-PVAF changes relative to the three initial Q matrices.}

\centering{

\centering
\begingroup\singlespacing\fontsize{9.5}{11.0}\selectfont
\setlength{\tabcolsep}{8pt}

\begin{tabular}{@{}llll@{}}
\toprule
Revision & Item & Attribute & Edit\\
\midrule
${Q_{\mathrm{i}1}\rightarrow Q_{\mathrm{r}1}}$ & M032132 & algebra & Addition\\
${Q_{\mathrm{i}1}\rightarrow Q_{\mathrm{r}1}}$ & M052216 & data and chance & Addition\\
${Q_{\mathrm{i}1}\rightarrow Q_{\mathrm{r}1}}$ & M042198A & data and chance & Addition\\
${Q_{\mathrm{i}1}\rightarrow Q_{\mathrm{r}1}}$ & M032132 & data and chance & Removal\\
${Q_{\mathrm{i}2}\rightarrow Q_{\mathrm{r}2}}$ & M052214 & S4 & Removal\\
${Q_{\mathrm{i}2}\rightarrow Q_{\mathrm{r}2}}$ & M032295 & S6 & Removal\\
${Q_{\mathrm{i}3}\rightarrow Q_{\mathrm{r}3}}$ & M052231 & algebra & Removal\\
${Q_{\mathrm{i}3}\rightarrow Q_{\mathrm{r}3}}$ & M042077 & algebra & Removal\\
${Q_{\mathrm{i}3}\rightarrow Q_{\mathrm{r}3}}$ & M032626 & number & Removal\\
${Q_{\mathrm{i}3}\rightarrow Q_{\mathrm{r}3}}$ & M032295 & number & Addition\\
${Q_{\mathrm{i}3}\rightarrow Q_{\mathrm{r}3}}$ & M042103 & number & Addition\\
\bottomrule
\end{tabular}
\endgroup

}

\end{table}%

\subsection{Diagnostic eligibility and response-model
comparisons}\label{diagnostic-eligibility-and-response-model-comparisons}

Two screens govern eligibility for a diagnostic claim, but they have
different status. Row coverage, at least one group association for every
item, is a prespecified practical eligibility rule for this application.
Separately, every proposed attribute must have an exclusive item so that
an identity-submatrix witness exists. Completeness is necessary for
strict identification in DINA; here it is a conservative application
rule, not a universal necessary condition for general G-DINA. The
separate general-RLCM structural check is reported below. Passing both
would still not prove full diagnostic identifiability
(\citeproc{ref-rupp2010diagnostic}{Rupp et al., 2010};
\citeproc{ref-xushang2018identifying}{Xu \& Shang, 2018}); failure of
either stops the diagnostic claim here, though it does not forbid a
clearly labeled exploratory response-model benchmark. The full
seven-group representation is primary. Any reduction after removing
shared-stem candidates requires independent expert adjudication and is
reported only as a screen, not as a final five-attribute Q-matrix.

Table \ref{tbl-screen} summarizes these application screens. In the
adopted Q, S6 and S7 have no exclusive indicators.

\begin{longtable}[]{@{}
  >{\raggedright\arraybackslash}p{(\linewidth - 12\tabcolsep) * \real{0.1000}}
  >{\raggedleft\arraybackslash}p{(\linewidth - 12\tabcolsep) * \real{0.1333}}
  >{\raggedleft\arraybackslash}p{(\linewidth - 12\tabcolsep) * \real{0.1333}}
  >{\centering\arraybackslash}p{(\linewidth - 12\tabcolsep) * \real{0.1667}}
  >{\raggedleft\arraybackslash}p{(\linewidth - 12\tabcolsep) * \real{0.1333}}
  >{\centering\arraybackslash}p{(\linewidth - 12\tabcolsep) * \real{0.1667}}
  >{\centering\arraybackslash}p{(\linewidth - 12\tabcolsep) * \real{0.1667}}@{}}
\caption{Diagnostic-feasibility
screen.}\label{tbl-screen}\tabularnewline
\toprule\noalign{}
\begin{minipage}[b]{\linewidth}\raggedright
Definition
\end{minipage} & \begin{minipage}[b]{\linewidth}\raggedleft
Reduced groups
\end{minipage} & \begin{minipage}[b]{\linewidth}\raggedleft
Reduced zero rows
\end{minipage} & \begin{minipage}[b]{\linewidth}\centering
Reduced identity witness
\end{minipage} & \begin{minipage}[b]{\linewidth}\raggedleft
Full zero rows
\end{minipage} & \begin{minipage}[b]{\linewidth}\centering
Full identity witness
\end{minipage} & \begin{minipage}[b]{\linewidth}\centering
Joint pass
\end{minipage} \\
\midrule\noalign{}
\endfirsthead
\toprule\noalign{}
\begin{minipage}[b]{\linewidth}\raggedright
Definition
\end{minipage} & \begin{minipage}[b]{\linewidth}\raggedleft
Reduced groups
\end{minipage} & \begin{minipage}[b]{\linewidth}\raggedleft
Reduced zero rows
\end{minipage} & \begin{minipage}[b]{\linewidth}\centering
Reduced identity witness
\end{minipage} & \begin{minipage}[b]{\linewidth}\raggedleft
Full zero rows
\end{minipage} & \begin{minipage}[b]{\linewidth}\centering
Full identity witness
\end{minipage} & \begin{minipage}[b]{\linewidth}\centering
Joint pass
\end{minipage} \\
\midrule\noalign{}
\endhead
\bottomrule\noalign{}
\endlastfoot
Depth-4 disjoint & 5 & 2 & No & 0 & No & No \\
Depth-4 overlap & 5 & 2 & Yes & 0 & No & No \\
Depth-2 overlap & 5 & 4 & No & 0 & No & No \\
Depth-2 disjoint & 6 & 0 & No & 0 & No & No \\
\end{longtable}

\begin{minipage}{\linewidth}
\footnotesize\textit{Note.} Reduced columns follow a statistical stem-concentration heuristic, not expert adjudication. Row coverage is practical; exclusive-item support is a model-dependent identity-submatrix witness. Passing both would not establish full identification.
\end{minipage}

The additional item-model comparison evaluates the initial 40-by-7
binary projection with HO-CDMs (\citeproc{ref-delatorre2011gdina}{{de la
Torre}, 2011}; \citeproc{ref-delatorredouglas2004higher}{{de la Torre \&
Douglas}, 2004}; \citeproc{ref-ma2020gdina}{{Ma \& de la Torre}, 2020}).
The general semantic column was excluded: it is present for every item
and supplies no diagnostic differentiation, whereas the higher-order
response trait governs attribute probabilities indirectly. The family
included G-DINA, A-CDM, and DINA. The estimator retains respondents with
more than one valid focal response, which excludes 14 of the 5,944
students who answered at least one, so every comparison used the same
5,930 respondents and 40 items, including a unidimensional 2PL reference
refitted to that exact matrix. The separate 5,944-person 2PL remains the
reference for the descriptive item crosswalk and grouped
parameter-prediction exercise.

Table \ref{tbl-hocdm-fit} combines diagnostics for the main comparison
with additional higher-order item-model comparisons. In both the
\(Q_{\mathrm{i}1}\) and \(Q_{\mathrm{r}1}\) HO fits, three of the four
attribute slopes reach the upper bound of 5; boundary estimates
therefore also affect the parsimonious official framework. The matched
2PL in Table \ref{tbl-r51-benchmark} has lower AIC and BIC than every
additional HO-CDM. Among CDMs, AIC favors the free-slope G-DINA
sensitivity whereas BIC favors A-CDM, so there is no criterion-invariant
internal winner. The unconstrained G-DINA item-response arms also show
boundary concentration and extensive nonmonotonicity. A
monotonic-constrained fixed-unit-slope fit removes the violations but
remains less competitive by both criteria.

\begin{table}

\caption{\label{tbl-hocdm-fit}Response-model diagnostics and additional model comparisons.}

\centering{

\centering
\begingroup\singlespacing
\fontsize{9.0}{10.5}\selectfont
\setlength{\tabcolsep}{4pt}

\label{tbl-r52-response-diagnostics}
\begin{tabular}{@{}llrrr@{}}
\toprule
\multicolumn{5}{@{}l}{\textit{Panel A. Diagnostics for the twelve CDM conditions}}\\[2pt]
Q & Dist. & \shortstack{Boundary\\items/cells} &
\shortstack{Nonmonotone\\items/decreases} & Pair RMS\\
\midrule
$Q_{\mathrm{i}1}$ & Sat. & 1/1 & 0/0 & .0201 \\
$Q_{\mathrm{i}1}$ & HO & 1/1 & 0/0 & .0206 \\
$Q_{\mathrm{r}1}$ & Sat. & 3/3 & 2/3 & .0198 \\
$Q_{\mathrm{r}1}$ & HO & 3/3 & 2/2 & .0204 \\
$Q_{\mathrm{i}2}$ & Sat. & 22/112 & 23/155 & .0190 \\
$Q_{\mathrm{i}2}$ & HO & 25/119 & 26/150 & .0193 \\
$Q_{\mathrm{r}2}$ & Sat. & 23/113 & 21/168 & .0196 \\
$Q_{\mathrm{r}2}$ & HO & 23/117 & 25/158 & .0192 \\
$Q_{\mathrm{i}3}$ & Sat. & 19/33 & 20/69 & .0193 \\
$Q_{\mathrm{i}3}$ & HO & 14/28 & 22/64 & .0188 \\
$Q_{\mathrm{r}3}$ & Sat. & 17/29 & 24/80 & .0197 \\
$Q_{\mathrm{r}3}$ & HO & 15/27 & 22/58 & .0187 \\
\bottomrule
\end{tabular}
\par\vspace{6pt}
\begin{tabular}{@{}lrrrrrr@{}}
\toprule
\multicolumn{7}{@{}l}{\textit{Panel B. Additional higher-order CDM comparisons}}\\[2pt]
Model & $p$ & log $L$ & AIC & BIC & \shortstack{Boundary\\items/cells} & \shortstack{Nonmonotone\\items/decr.}\\
\midrule
G-DINA-C & 297 & -30,872.48 & 62,338.96 & 64,325.23 & 24/79 & 19/129\\
A-CDM-C & 140 & -31,059.01 & 62,398.01 & 63,334.30 & 18/22 & 4/50\\
DINA-C & 87 & -31,382.15 & 62,938.31 & 63,520.14 & 8/12 & 0/0\\
G-DINA-F & 304 & -30,589.01 & 61,786.02 & 63,819.11 & 25/119 & 26/150\\
G-DINA-M & 297 & -30,919.17 & 62,432.34 & 64,418.61 & 12/23 & 0/0\\
\bottomrule
\end{tabular}
\par\vspace{3pt}\parbox{\textwidth}{\footnotesize Note. $N=5{,}930$ and 40 items throughout. Boundary cells have reduced-pattern probabilities below .001 or above .999. Pair RMS is the root mean square observed-minus-implied joint-correct probability over 322 coobserved pairs. All CDM fits converge with finite estimates, probabilities, and item-parameter SEs; higher-order structural SEs were not estimated. Panel B uses $Q_{\mathrm{i}2}$: C = fixed-unit attribute slopes, F = free positive attribute slopes, and M = fixed-unit attribute slopes with monotonic item-response constraints. The F row repeats the $Q_{\mathrm{i}2}$ HO fit in the main comparison. The 2PL reference is in Table \ref{tbl-r51-benchmark}; the CDM-specific checks here were not recorded for it.}
\endgroup

}

\end{table}%

In the restricted-model comparison, an absolute-fit statistic could not
be obtained: the package call returned ``missing value where TRUE/FALSE
needed.'\,' We record that error without attributing a cause.

The auxiliary checks in Table \ref{tbl-hocdm-sensitivity} prevent a
simple \emph{useful} versus \emph{useless} verdict. The adopted G-DINA
placement is better than all 10 row-permuted matrices that preserve its
density and column structure, although the plus-one tail resolution is
only \(1/11=.091\). The adopted placement includes 16 official-content
anchors among its 28 imposed group memberships, as well as added
semantic markers. Its advantage therefore concerns this combined item
placement rather than solely discovered semantic structure, and it is
insufficient for the joint projection--model combination to earn its
complexity over the 2PL.

All four text-based projections are computationally estimable after the
declared depth-two-overlap seed escalation. AIC favors the depth-two
disjoint projection, whereas BIC favors depth-four overlap, so the
response results do not reselect the adopted text-side definition. DINA
produces 112 distinct ideal-response signatures among 128 profiles. The
adopted Q satisfies structural Conditions D and E of Theorem 4 in Gu and
Xu's general-RLCM analysis (\citeproc{ref-guxu2019q}{Gu \& Xu, 2019}):
two disjoint square item submatrices have diagonal ones after ordering,
and the remaining items cover every attribute. The two item blocks and
residual coverage are deposited. This is a structural check relevant to
generic joint identification of \((Q,\Theta,p)\) up to column
permutation under the theorem's model assumptions. Those assumptions
include positive class proportions, local independence, and the
specified capable/incapable response-probability restrictions. The
theorem uses the full joint response distribution. The relevant
determinant and distinctness conditions at the fitted parameter values,
extension to the higher-order submodel, and sufficiency of the observed
booklet marginals have not been verified. Consequently, this witness is
not a certification of the fitted boundary-heavy response model or of
diagnosis under the sparse design. The higher-order response score
correlates .951 with 2PL ability, which compares two response-derived
rankings rather than testing semantic G. At item level, the G-DINA
discrimination index and semantic G remain weakly related (\(r=-.198\),
permutation \(p=.213\)).

These checks do not establish diagnosis under the observed design. Table
\ref{tbl-hocdm-sensitivity} reports the complete mapping comparisons; no
individual attribute classifications or mastery rates are inferred.

\begin{table}

\caption{\label{tbl-hocdm-sensitivity}Higher-order G-DINA mapping sensitivities.}

\centering{

\centering
\begingroup\singlespacing\small
\setlength{\tabcolsep}{4.0pt}
\renewcommand{\arraystretch}{1.00}

\begin{tabular}{@{}lrrrr@{}}
\toprule
\multicolumn{5}{@{}l}{\textit{Panel A. Four text-based projections}}\\[2pt]
Definition & $p$ & log $L$ & AIC & BIC\\
\midrule
Depth-4 disjoint & 297 & -30,872.48 & 62,338.96 & 64,325.23\\
Depth-4 overlap & 263 & -30,860.22 & 62,246.43 & 64,005.32\\
Depth-2 overlap, escalated & 293 & -30,891.72 & 62,369.44 & 64,328.96\\
Depth-2 disjoint & 325 & -30,792.76 & 62,235.52 & 64,409.05\\
\midrule
\multicolumn{5}{@{}l}{\textit{Panel B. Density-matched placement null}}\\[2pt]
Object & & log $L$ & AIC & BIC\\
\midrule
Adopted placement & & -30,872.48 & 62,338.96 & 64,325.23\\
Permuted median & & -31,006.93 & 62,607.87 & 64,594.14\\
Permuted range & & \shortstack{-31,102.31 to\\-30,897.16} & \shortstack{62,388.31--\\62,798.61} & \shortstack{64,374.58--\\64,784.88}\\
\bottomrule
\end{tabular}
\vspace{2pt}\parbox{.94\textwidth}{\footnotesize Note. $N=5{,}930$, 40 items, and fixed-unit-slope higher-order G-DINA throughout. All four projections were computationally estimable. In Panel A, AIC favors depth-2 disjoint and BIC favors depth-4 overlap; responses therefore do not select among the text-based definitions. The initial depth-2 overlap fit reached the 2,000-iteration cap. Ten external seeds (123456--123465) were then run with a 5,000-iteration cap and the three internal starts enforced by \texttt{GDINA} 2.9.12. All converged; the best-likelihood run (seed 123457; 2,590 iterations) is shown. These are separate seed-based fits, each with three internal starts. In Panel B, row permutations preserve density, column margins, and the row-weight multiset while disrupting item placement. The adopted placement is favorable in all 10 permutations; the plus-one descriptive tail is $1/11=.091$.}
\endgroup

}

\end{table}%

\subsection{Residual dependence}\label{residual-dependence}

Signed group-loading geometry is compared with Rasch \(Q_3\) over the
item pairs observable under the booklet design, with within-bundle pairs
excluded. The semantic predictor is the inner product of each pair's
signed group-loading vectors; the general factor is excluded. A
reference set contains 200 simulated \(Q_3\) matrices. Each originates
from Rasch responses generated using standard normal abilities,
published item difficulties, and the observed missingness mask, then
processed through the same penalized joint Rasch-to-\(Q_3\) procedure.
For each item set, the calibrated \(z\) is the observed Pearson
geometry--\(Q_3\) correlation minus its null-bank mean, divided by its
null-bank standard deviation. These are calibration diagnostics rather
than independent-pair significance tests. The fitted Rasch model
converged.

Table \ref{tbl-q3} shows a weak association between the continuous
semantic geometry and response residual dependence. Removing shared-stem
bundles increases the association, but it remains small. The text
predictor is held fixed throughout these comparisons.

\begin{table}

\caption{\label{tbl-q3}Residual-geometry associations.}

\centering{

\centering
\begingroup\singlespacing\small

\begin{tabular}{@{}lrrrr@{}}
\toprule
Item set & Item pairs & Pearson $r$ & Spearman $\rho$ & Null-calibrated $z$ \\
\midrule
All 40 items & 316 & .035 & .031 & 1.93 \\
Exclude M042169 & 262 & .041 & .029 & 2.63 \\
Exclude M042198 & 262 & .079 & .074 & 2.46 \\
Exclude both bundles & 217 & .093 & .071 & 2.99 \\
\bottomrule
\end{tabular}
\vspace{2pt}\parbox{.94\textwidth}{\footnotesize Note. The all-item row contains the 316 $Q_3$-estimable pairs remaining after the six within-bundle pairs are excluded; deletion rows subset the same predictor without refitting.}
\endgroup

}

\end{table}%

\subsection{Grouped item-parameter
prediction}\label{grouped-item-parameter-prediction}

The ancillary prediction check asks whether the unfactored text
representations predict item parameters. Nested ridge regressions
predicted 2PL difficulty and log-discrimination directly from either the
1,864 contextual scores or the 4,096 raw embedding coordinates. Outer
and inner folds kept the three parts of each shared stem together; 20
fixed grouped 10-fold assignments quantified split variation. The
performance statistic is held-out predictive \(R^2_{\mathrm{pred}}\),
\(1-\mathrm{SSE}_{\mathrm{OOF}}/\mathrm{SST}\), where
\(\mathrm{SSE}_{\mathrm{OOF}}\) is the out-of-fold sum of squared errors
and \(\mathrm{SST}\) is the total sum of squares. A negative value
therefore means that the held-out predictions are worse than the
overall-mean reference.

Only contextual-score prediction of difficulty has a positive mean in
Table \ref{tbl-parameter-prediction}, and its gain is small. The raw
embedding does worse than the mean reference for difficulty, and both
representations do worse for log-discrimination.
Leave-one-base-item-group-out results are similar. This bank therefore
contains little representation-to-parameter signal recoverable by the
evaluated linear regularization. The result does not distinguish
insufficient item count from a weak, diffuse, nonlinear, or
representation-specific relationship.

For parameter-precision context, the ratios
\(1-\operatorname{mean}(\mathrm{SE}^2)/\operatorname{var}(\widehat\theta)\)
are .9819 for difficulty and .9227 for log-discrimination. Under an
unbiased, independent calibration-error interpretation with correctly
specified standard errors, they approximate the share of between-item
variation beyond estimation error. Thus reported calibration SEs alone
appear unlikely to explain the weak prediction. These model-based ratios
do not rule out misspecification or other noise, and they leave the
small item-level training sample as a limitation.

\begingroup\footnotesize

\begin{longtable}[]{@{}
  >{\raggedright\arraybackslash}p{(\linewidth - 10\tabcolsep) * \real{0.1852}}
  >{\raggedright\arraybackslash}p{(\linewidth - 10\tabcolsep) * \real{0.1111}}
  >{\raggedleft\arraybackslash}p{(\linewidth - 10\tabcolsep) * \real{0.1852}}
  >{\raggedleft\arraybackslash}p{(\linewidth - 10\tabcolsep) * \real{0.1852}}
  >{\raggedleft\arraybackslash}p{(\linewidth - 10\tabcolsep) * \real{0.1481}}
  >{\raggedleft\arraybackslash}p{(\linewidth - 10\tabcolsep) * \real{0.1852}}@{}}
\caption{Embedding-to-parameter
prediction.}\label{tbl-parameter-prediction}\tabularnewline
\toprule\noalign{}
\begin{minipage}[b]{\linewidth}\raggedright
Representation
\end{minipage} & \begin{minipage}[b]{\linewidth}\raggedright
Target
\end{minipage} & \begin{minipage}[b]{\linewidth}\raggedleft
Mean \(R^2_{\mathrm{pred}}\) (SD)
\end{minipage} & \begin{minipage}[b]{\linewidth}\raggedleft
Range
\end{minipage} & \begin{minipage}[b]{\linewidth}\raggedleft
Positive repeats
\end{minipage} & \begin{minipage}[b]{\linewidth}\raggedleft
Leave-one-group-out \(R^2_{\mathrm{pred}}\)
\end{minipage} \\
\midrule\noalign{}
\endfirsthead
\toprule\noalign{}
\begin{minipage}[b]{\linewidth}\raggedright
Representation
\end{minipage} & \begin{minipage}[b]{\linewidth}\raggedright
Target
\end{minipage} & \begin{minipage}[b]{\linewidth}\raggedleft
Mean \(R^2_{\mathrm{pred}}\) (SD)
\end{minipage} & \begin{minipage}[b]{\linewidth}\raggedleft
Range
\end{minipage} & \begin{minipage}[b]{\linewidth}\raggedleft
Positive repeats
\end{minipage} & \begin{minipage}[b]{\linewidth}\raggedleft
Leave-one-group-out \(R^2_{\mathrm{pred}}\)
\end{minipage} \\
\midrule\noalign{}
\endhead
\bottomrule\noalign{}
\endlastfoot
Contextual scores & 2PL \(b\) & .025 (.032) & -.074 to .058 & 17/20 &
.030 \\
Raw item embedding & 2PL \(b\) & -.094 (.039) & -.195 to -.021 & 0/20 &
-.076 \\
Contextual scores & 2PL \(\log(a)\) & -.094 (.054) & -.255 to -.026 &
0/20 & -.071 \\
Raw item embedding & 2PL \(\log(a)\) & -.141 (.061) & -.261 to -.059 &
0/20 & -.170 \\
\end{longtable}

\begin{minipage}{\linewidth}
\footnotesize\textit{Note.} Twenty fixed grouped 10-fold assignments with inner tuning. Leave-one-group-out holds out one base-item group. Response parameters never construct or select a text-side factor or mapping.
\end{minipage}

\endgroup

\end{document}